%% file: sample-sigconf.tex
\documentclass[sigconf]{acmart}
\usepackage{subcaption}
\usepackage{diagbox} 
\usepackage{stmaryrd}
\usepackage[hyphenbreaks]{breakurl}
\usepackage{hyperref}
\hypersetup{
	colorlinks=true,
	linkcolor=blue,
	filecolor=magenta,      
	urlcolor=cyan,
	pdftitle={Overleaf Example},
	pdfpagemode=FullScreen,
}

\AtBeginDocument{%
  \providecommand\BibTeX{{%
    \normalfont B\kern-0.5em{\scshape i\kern-0.25em b}\kern-0.8em\TeX}}}

\setcopyright{acmcopyright}

\copyrightyear{2021} 
\acmYear{2021} 
\setcopyright{acmcopyright}\acmConference[MM '21]{Proceedings of the 29th ACM International Conference on Multimedia}{October 20--24, 2021}{Virtual Event, China}
\acmBooktitle{Proceedings of the 29th ACM International Conference on Multimedia (MM '21), October 20--24, 2021, Virtual Event, China}
\acmPrice{15.00}
\acmDOI{10.1145/3474085.3475272}
\acmISBN{978-1-4503-8651-7/21/10}

\acmSubmissionID{568}

\begin{document}

\title{Token Shift Transformer for Video Classification}
\fancyhead{}
\author{Hao Zhang}
\affiliation{%
  \institution{City University of Hong Kong}
  \city{Hong Kong SAR}
  \country{China}
}
\email{zhanghaoinf@gmail.com}

\author{Yanbin Hao}
\authornote{Yanbin Hao is the corresponding author.}
\affiliation{%
  \institution{University of Science and Technology of China}
  \city{Hefei}
  \country{China}}
\email{haoyanbin@hotmail.com}

\author{Chong-Wah Ngo}
\affiliation{%
  \institution{Singapore Management University}
  \country{Singapore}
}
\email{cwngo@smu.edu.sg}

\renewcommand{\shortauthors}{Trovato and Tobin, et al.}

\begin{abstract}
 Transformer achieves remarkable successes in understanding 1 and 2-dimensional signals (e.g., NLP and Image Content Understanding). As a potential alternative to convolutional neural networks, it shares merits of strong interpretability, high discriminative power on hyper-scale data, and flexibility in processing varying length inputs.  However, its encoders naturally contain computational intensive operations such as pair-wise self-attention, incurring heavy computational burden when being applied on the complex 3-dimen\\sional video signals.

  This paper presents \textbf{Token Shift Module} (i.e., TokShift), a novel, zero-parameter, zero-FLOPs operator, for modeling temporal relations within each transformer encoder. Specifically, the TokShift barely temporally shifts partial [\textit{Class}] \textbf{token} features back-and-forth across adjacent frames. Then, we densely plug the module into each encoder of a plain 2D vision transformer for learning 3D video representation. It is worth noticing that our TokShift transformer is a pure convolutional-free video transformer pilot with computational efficiency for video understanding. Experiments on standard benchmarks verify its robustness, effectiveness, and efficiency. Particularly, with input clips of 8/12 frames, the TokShift transformer achieves SOTA precision: 79.83\%/80.40\% on the Kinetics-400, 66.56\% on EGTEA-Gaze+, and 96.80\% on UCF-101 datasets, comparable or better than existing SOTA convolutional counterparts. Our code is open-sourced in: \sloppy \href{https://github.com/VideoNetworks/TokShift-Transformer}{\textit{\color{magenta}{https://github.com/VideoNetworks/TokShift-Transformer}}}.
\end{abstract}

\ccsdesc[500]{Computing methodologies~Activity recognition and understanding}



\keywords{Video classification; Transformer; Shift; Self-attention}


\maketitle

\input{part1_intro}
\input{part2_related}
\input{part3_method}
\input{part4_exp}

\section{Conclusion}
 We propose a zero-parameter, zero-FLOPs TokShift operator for constructing a pure convolutional-free video transformer. Specifically, our TokShift-xfmr alleviates intensive pair-wise distance calculations of the Spatio-temporal attention, maintains the same complexity as a common 2D ViT, while achieving better or comparable performance (80.40\%) as 3D-CNN SOTAs. More importantly, TokShift firstly conducts temporal modeling on the global token of video transformers. Since global [\textit{Class}] token is aggregated by weighted-sum all spatial patches' embeddings and reflects global visual content. Hence, modeling temporal interactions across frames can be done through the tokens. Besides, we still face an inherent computational burden of attention, within the 2D ViT, especially in processing long-length video. We leave computational optimization of video transformer as future works. Finally, with the help of our TokShift-xfmr, we can partially answer the prophecy of the quote on video: ``\textit{A 10 seconds video is worthy of 3,152 visual words}''.
\section{Acknowledgments}
I thank Dr. Jingjing Chen and Dr. Bin Zhu for inspiring discussion and encouragement.


\vfill\eject
\bibliographystyle{ACM-Reference-Format}
\bibliography{sample-base}


\end{document}

%% file: part1_intro.tex
\section{Introduction}
``\textit{If a picture is worth a thousand words, is a video worth a million?}''  \cite{yadav2011if}. This quote basically predicts that image and video can be potentially interpreted as linguistical sentences, except that videos
contain richer information than images. The recent progress of extending linguistical-style, convolutional-free transformers \cite{dosovitskiy2020image} on visual content understanding successfully verifies the quote's prior half, whereas the latter half for video remains an open hypothesis. This paper studies the efficient and effective way of applying similar transformers for video understanding and answer the quote's prophecy for videos.
\begin{figure}
    \centering
    \begin{subfigure}[t]{0.17\textwidth}
        \includegraphics[width=\textwidth]{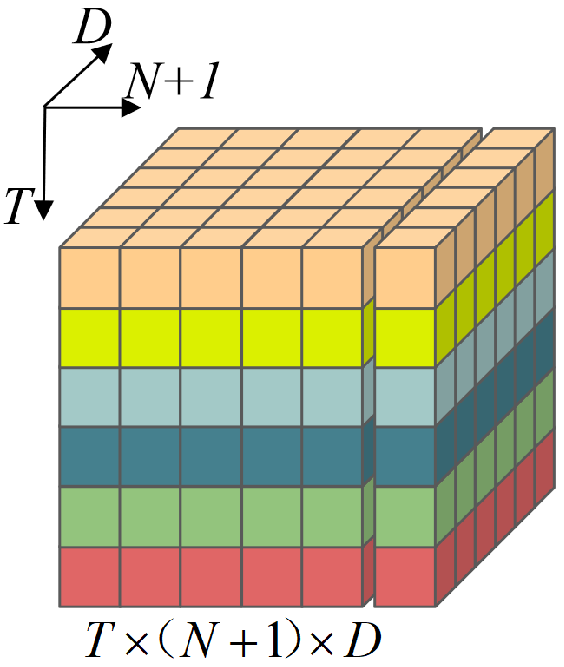}
        \caption{Video Tensor}
        \label{fig:tensor}
    \end{subfigure}
    \begin{subfigure}[t]{0.17\textwidth}
        \includegraphics[width=\textwidth]{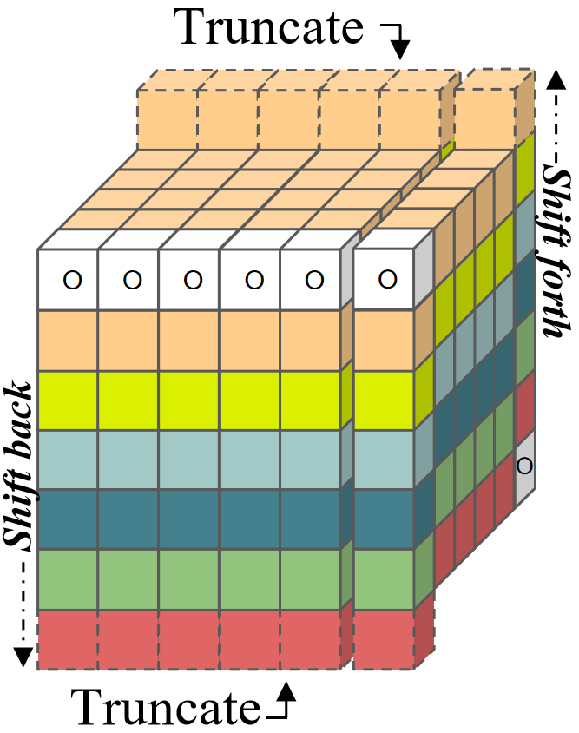}
        \caption{Temporal Shift}
        \label{fig:tensor_shift}
    \end{subfigure}
    \\
    ~ 
    \begin{subfigure}[t]{0.17\textwidth}
        \includegraphics[width=\textwidth]{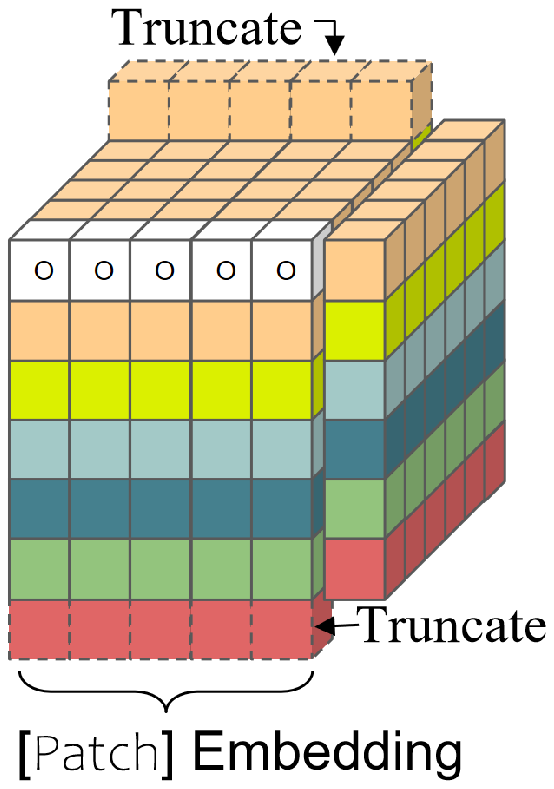}
        \caption{Patch Shift}
        \label{fig:tensor_patch}
    \end{subfigure}
    \begin{subfigure}[t]{0.17\textwidth}
        \includegraphics[width=\textwidth]{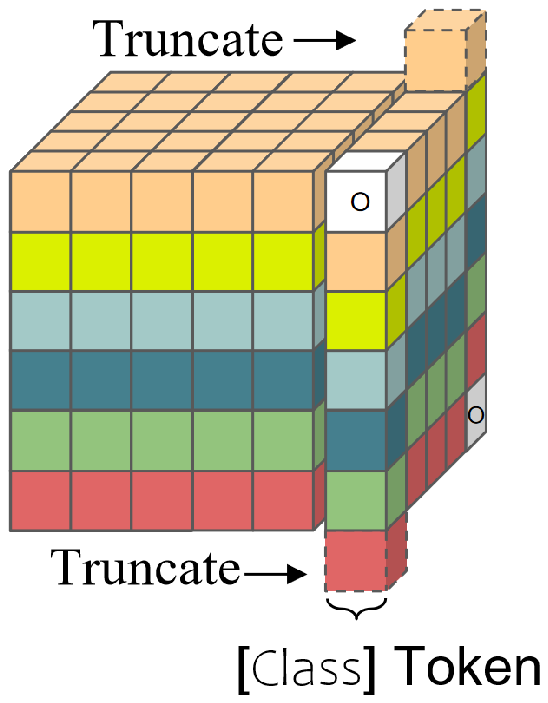}
        \caption{Token Shift}
        \label{fig:tensor_class}
    \end{subfigure}
    \vspace{-0.4cm}
    \caption{Types of \textbf{Shift} for video transformer. A video embedding contains two types of ``words'': $N$ [\textit{Patch}] + $1$ [\textit{Class}] Token. Hence, the Shift can be applied on: (a) \textit{Neither}, (b) \textit{Both}, (c) \textit{Patch}, and (d) [\textit{Class}] \textit{Token} along the temporal axis. ``$\boxcircle$'' indicates padding zeros.}\label{fig:all_shift}
    \vspace{-0.5cm}
\end{figure}

As a potential alternative for convolutional neural networks, the transformer achieves remarkable progress in both NLP \cite{brown2020language,beltagy2020longformer,vaswani2017attention,devlin2018bert,kitaev2020reformer} and vision tasks \cite{dosovitskiy2020image,touvron2020deit,carion2020end,wang2020end,wang2021m2tr}. Particularly, language and vision transformers show several merits over their CNN counterparts, such as good interpretability (e.g., attention to highlight core parts), unsaturated discriminatory power scalable with hyper-scale data (e.g., million-scale language corpus\cite{vaswani2017attention, devlin2018bert} or billion-scale images\cite{dosovitskiy2020image}), and flexibility in processing varying length inputs \cite{kitaev2020reformer, beltagy2020longformer}.

However, unlike signal understanding on 1 and 2-dimensional data (i.e., NLP \& static vision), the application of transformers on 3-dimensional video signals is challenging. Specifically, each encoder of a transformer naturally contains heavy computations such as pair-wise self-attention; meanwhile, a video has a longer sequential representation ($\pmb{z}_{v}\in \mathbb{R}^{T\times(N+1)\times D}$) than an image ($\pmb{z}_{i}\in \mathbb{R}^{(N+1)\times D}$) due to an extra temporal axis. Consequently, directly applying general transformers on flattened spatio-temporal video sequences will introduce an exponential explosion of computations (e.g., $T^2\cdot\frac{(N+1)^2}{2}$ pair-wise distance calculations\footnote{$T$, $N$, $1$, $D$ denote time-stamps, spatial-patches, [Class] token and feature-dim}) in the training and inference phases.

To tackle this, we propose \textbf{Token Shift Module} (i.e., TokShift), a novel zero-parameter, zero-FLOPs operator, for modeling temporal relations within each video encoder. Specifically, the TokShift module barely temporally shifts partial [\textit{Class}] \textbf{token} features back-and-forth (Figure (\ref{fig:tensor_class})) across frames. Then, we densely plug the TokShift module into each encoder of a plain 2D vision transformer for learning 3D video representation. 

Our TokShift is partially inspired by the success of spatial \cite{chen2019all,wu2018shift,you2020shiftaddnet} and temporal \cite{lin2019tsm,fan2020rubiksnet} shift operators for efficiency optimization on CNNs, but bears its own uniqueness. Specifically, on CNNs, spatial/temporal shift is uniformly applied across all spatial receptive fields on a feature-map. Follow this imitation, a copycat of the Temporal Shift Module for transformers is shown in Figure (\ref{fig:tensor_shift}), where the shift is imposed on all patches and [\textit{Class}] token features. However, we experimentally verify that, rather than temporally shifting features of all reception fields, just shift [\textit{Class}] \textbf{token} feature is sufficient. As in Figure (\ref{fig:tensor_shift}-\ref{fig:tensor_class}), our TokShift introduces minimum modifications on original feature $\pmb{z}_{v}$, and gets the most improvement among all shift variants. More importantly, equipped with the TokShift, a video transformer only needs to handle $T\cdot\frac{(N+1)^2}{2}$ pair-wise calculations. Finally, we verify that the shift operator is generalizable on transformers as it does on CNNs.

We try to intuitively explain the efficacy of the TokShift by drawing an analogy between \textit{textual} description and \textit{visual} feature vector of each frame. Specifically, the text describing a frame usually follow a similar linguistical form: ``verb'' + ``noun'' (e.g., ``\textit{walking the dog}'' in  Figure (\ref{fig:attention})), where ``verb'' is dynamically correlated with neighboring frames and ``noun'' are static. Correspondingly, the TokShift exchanges partial visual features of a frame (in the form of a token vector) temporally back-and-forth for motion capturing while keeping the rest for static semantics modeling.

We conduct extensive experiments of the TokShift-xfmr on standard benchmarks, such as Kinetics-400 \cite{carreira2017quo}, EGTEA-Gaze+ \cite{li2018eye} and UCF-101 \cite{soomro2012ucf101} datasets.~Experiments demonstrate that with the TokShift, video transformers could perform comparable or better than the best 3D-CNNs, reaching SOTA performance. Our contributions are summarized below.

\begin{itemize}
\item \textbf{Transformer for video classification.} Our TokShift-xfmr is an efficient pure convolutional-free transformer pilot applied for video classification. We repeat the successes of transformers on 3D videos as on 1/2D texts/images, promoting applications of transformers on different domains.
\item \textbf{Efficient TokShift for transformer encoder.} We specifically design a novel zero-parameter, zero-FLOPs TokShift operator for the encoder. By merely manipulating global frame representation (i.e., [\textit{Class}] token vector) via shift operations, we can reduce computations of pair-wise attention in dealing with sequential flattening video representations within transformers.
\item \textbf{State-of-the-art performances.} We test the TokShift-xfmr on several standard benchmarks. Experiments show that the TokShift-xfmr achieves SOTA top-1 accuracy 79.83\%/80.40\% (8/12f) on Kinetics-400, 66.56\% (8f) on EGTEA-Gaze+, and 96.80\% (8f) on UCF-101 datasets, comparable or better than the best 3D-CNNs available (TSM, SlowFast-R101-NL, X3D-XXL).
\end{itemize}

%% file: part2_related.tex
\section{Related Works}
The TokShift-xfmr is closely relevant to research areas, including 3D-CNNs for video classification, shift strategies for convolutional optimization, and 1/2D transformers for NLP/vision understanding; we separately elaborate the related literature for each area as below. 

\textbf{3D-CNNs for video classifications.} Convolutional neural networks stepped into a prosperous age in the past few years \cite{krizhevsky2012imagenet,simonyan2014very,he2016deep,chen2016deep,zhang2018fine,zhang2016object,su2020video,song2021spatial,wu2020interpretable,zhu2020cross,wei2020heuristic,zhao2020clean}. Consequently, the 3D-CNN \cite{tran2015learning,simonyan2014two,carreira2017quo,qiu2017learning,xie2018rethinking,feichtenhofer2019slowfast,feichtenhofer2020x3d,hao2020person,hao2020compact,qiu2019learning} has become a de-facto standard for video content understanding. Specifically, C3D \cite{tran2015learning} firstly inflates the convolutional kernel from 2D to 3D, to facilitate temporal modeling. For motion capturing, the two-stream network \cite{simonyan2014two} fuses two CNNs trained on optical-flow with RGB modalities. Furthermore, I3D \cite{carreira2017quo} deepens 3D-CNNs by inflating 3D kernels on inception-net, accompanied with a large-scale Kinetics dataset. To reduce computational overhead, Pseudo-3D \cite{qiu2017learning} decomposes a 3D Spatio-temporal convolution into 
Spatial-2D + Temporal-1D form. This decomposition greatly balances computations, parameters, and performances, hence gain its popularity in later 3D-CNNs. S3D-G \cite{xie2018rethinking} scales channel elements with attention yielded by global feature. Non-local networks \cite{wang2018non} introduces self-attention on top of CNNs; furthermore, CBA-QSA CNN \cite{hao2020compact} extends self-attention with compact bilinear mapping for fine-grained action classification. SlowFast \cite{feichtenhofer2019slowfast} and X3D \cite{feichtenhofer2020x3d} are currently best two 3D-CNNs. The former designs dual-paths CNNs, where each path receives input clip of slow/fast sampling rate; the latter presents an efficient strategy to search for optimal hyperparameters (e.g., spatial/temporal resolutions, channels, depth, etc.) on a template network (i.e., X2D).

\begin{figure*}[h!t]
\vspace{-0.1cm}
    \begin{center}
    \includegraphics[width=0.85\textwidth]{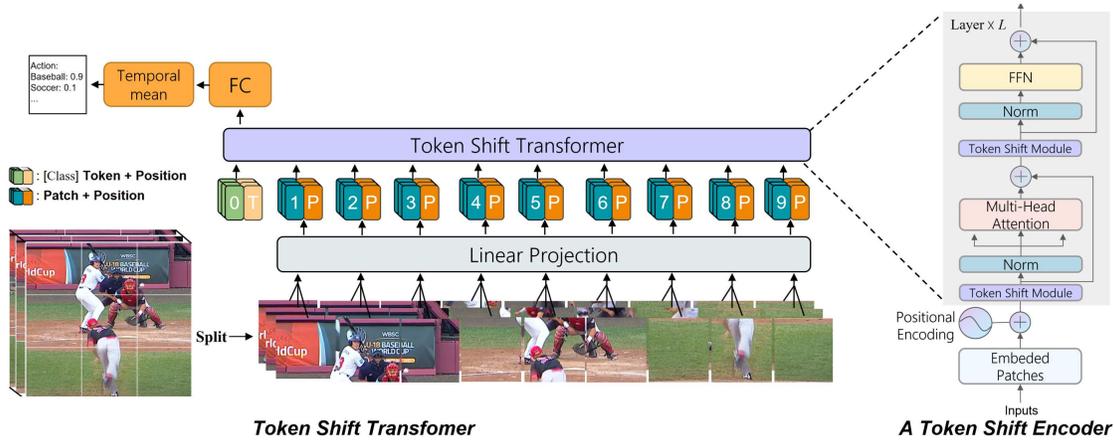}
    \end{center}
    \vspace{-0.5cm}
       \caption{\textbf{Token Shift Transformer}: the input video is divided into multiple patch clips, according to the rigid spatial grids. Patch clips are linear-projected, added with positional embeddings ($\pmb{E}_{pos}$), and then concatenated with a [\textit{Class}] token tensor ($\pmb{c}_0$) to form video embedding ($\pmb{z}_0$). The $\pmb{z}_0$ is fed into $L$ layers of identical TokShift encoders for learning video representation.}
    \label{fig:overview}
    \vspace{-0.5cm}
\end{figure*}

\textbf{Spatial/temporal shift for efficient CNNs.} Mobile computing demands more efficiency than cloud computing. Thereby, the Shift \cite{wu2018shift,you2020shiftaddnet,chen2019all,lin2019tsm,fan2020rubiksnet}, a zero-parameter, zero-FLOPs operator for local feature aggregation, is proposed to reduce the complexities of CNNs. Specifically, Shift + 1D/2D convolution can lossily approximate a 2D/3D convolution. For example, in \cite{wu2018shift}, Wu replaces spatial convolution of 3$\times$3 kernel  with Shift + 1$\times$1 kernel for vision understanding. To further improve efficiency, ShiftAddNet \cite{you2020shiftaddnet} introduces Shift + Add, eliminating all multiplications. For videos, Lin \cite{lin2019tsm} constructs TSM with Temporal Shift + spatial convolutions; RubiksNet \cite{fan2020rubiksnet} further removes spatial convolutions and introduces a learnable spatio-temporal shift operator. 

\textbf{Language \& vision transformer.} Firstly proposed in \cite{vaswani2017attention}, the transformer relies on residual attention \& feedforward (FFN) in feature learning and outperforms CNNs to become the de-facto standard in NLP tasks \cite{devlin2018bert,brown2020language}.  To tackle inputs of different/long length, Reformer \cite{kitaev2020reformer} and Longformer \cite{beltagy2020longformer} introduce optimizations on attention calculations. Recently, researchers repeat this success in vision tasks, including image classification \cite{dosovitskiy2020image,touvron2020deit}, object detection \cite{carion2020end}, and segmentations \cite{wang2020end}. Specifically, Dosovitskiy \cite{dosovitskiy2020image} proposes the first pure convolutional-free vision transformer (i.e., ViT) for image classification. To tackle the generalization problem caused by insufficient data, they equip the pre-training phase with a hyper-scale internal dataset (JFT-300M). In \cite{touvron2020deit}, DeiT bypasses the necessity of pre-training transformer on hyper-scale data through distillation. They introduce the token distillation for transformers to teach ``student'' transformer from various ``teachers'' (i.e., convnet or deeper transformer). For object detection and segmentation, as the transformer contains positional encodings under the encoder/decoder mechanism, it can be naturally extended to regress coordinates \cite{carion2020end} and masks \cite{wang2020end}. Overall, transformers are strong alternatives to CNNs and share merits like good interpretability, unsaturated discriminative power and flexibility on input length.

%% file: part3_method.tex
\section{Method}
\begin{figure*}
    \centering
    \scriptsize
    \begin{subfigure}[t]{0.186\textwidth}
        \includegraphics[width=\textwidth]{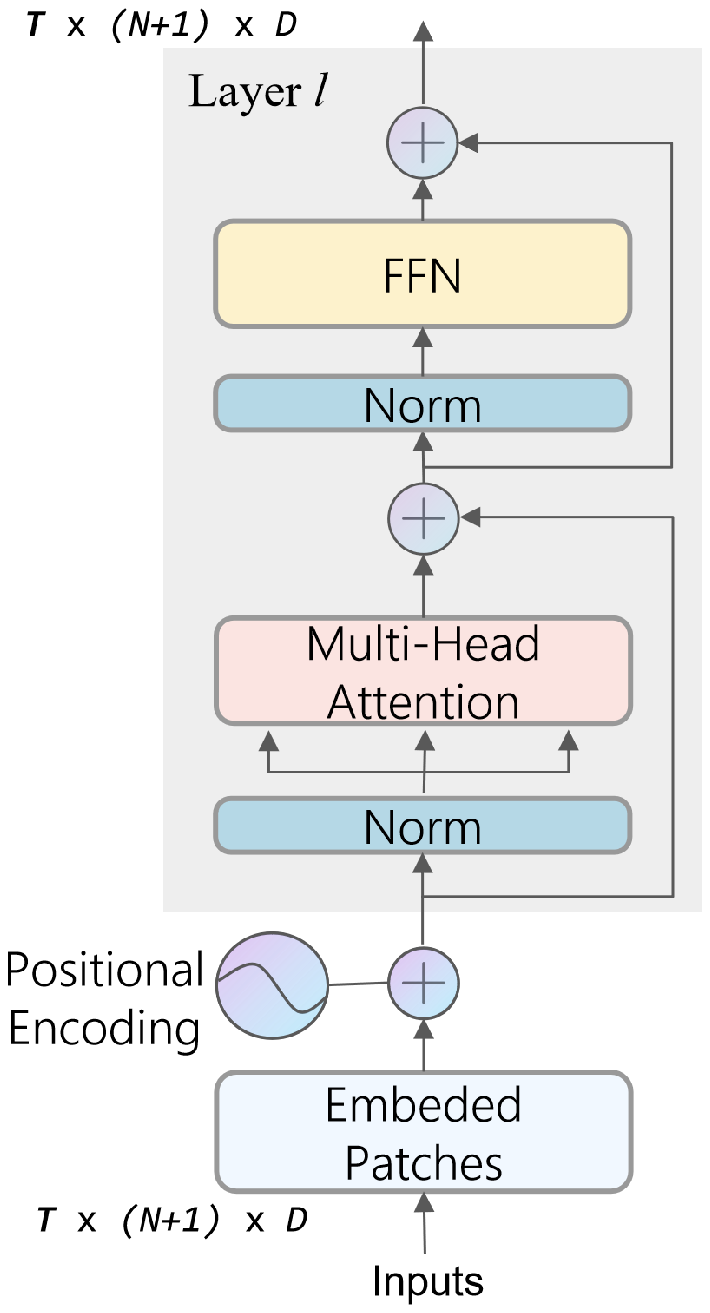}
        \caption{ViT (video)}
        \label{fig:vit}
    \end{subfigure}
        \begin{subfigure}[t]{0.186\textwidth}
        \includegraphics[width=\textwidth]{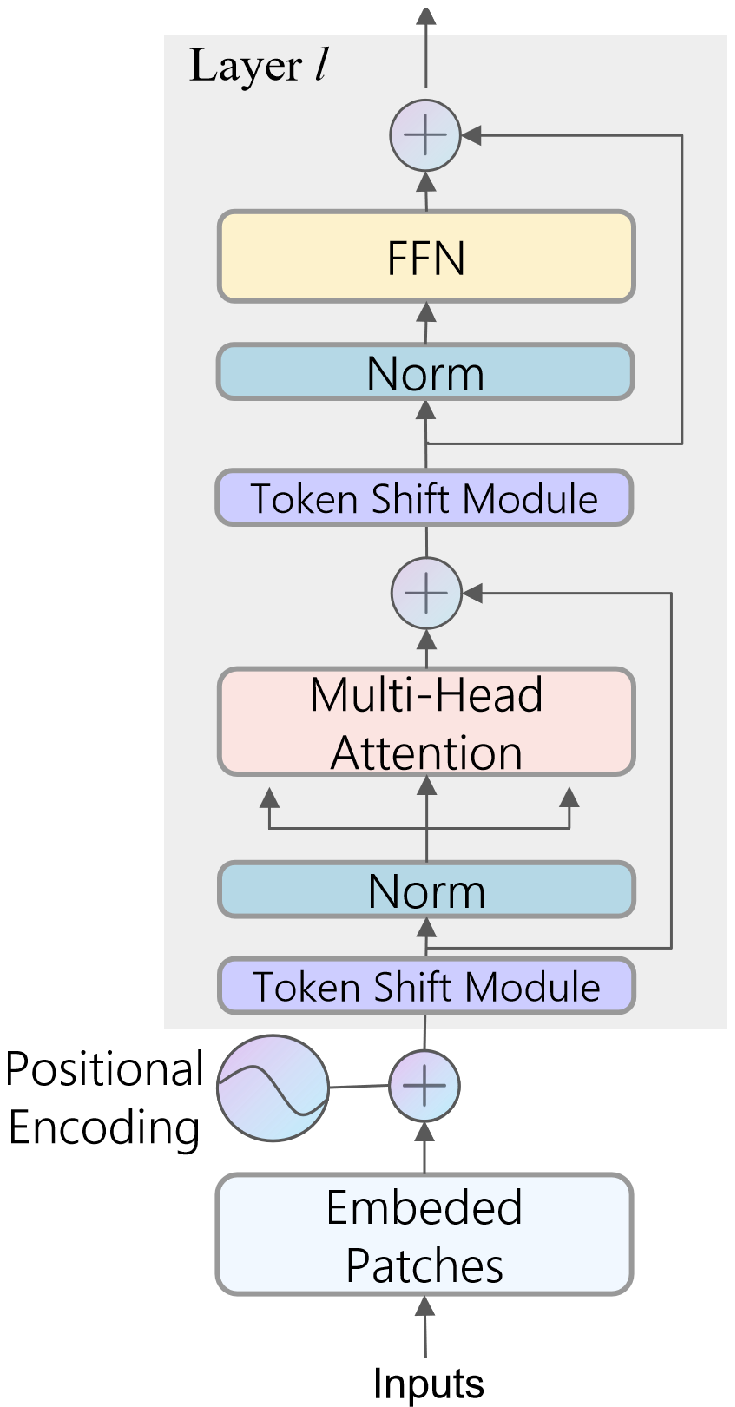}
        \caption{TokShift}
        \label{fig:vist_d}
    \end{subfigure}
    \begin{subfigure}[t]{0.186\textwidth}
        \includegraphics[width=\textwidth]{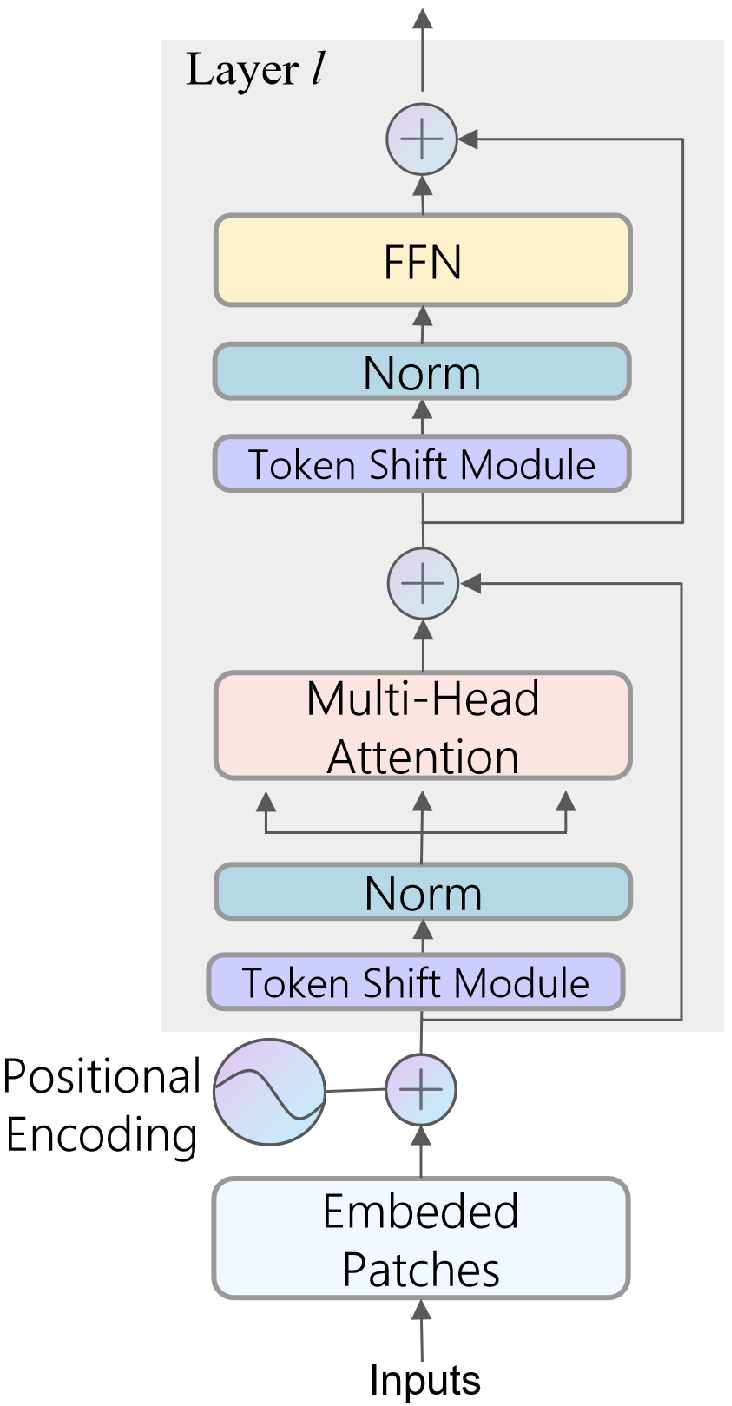}
        \caption{TokShift-\textit{A}}
        \label{fig:vist_a}
    \end{subfigure}
    \begin{subfigure}[t]{0.186\textwidth}
        \includegraphics[width=\textwidth]{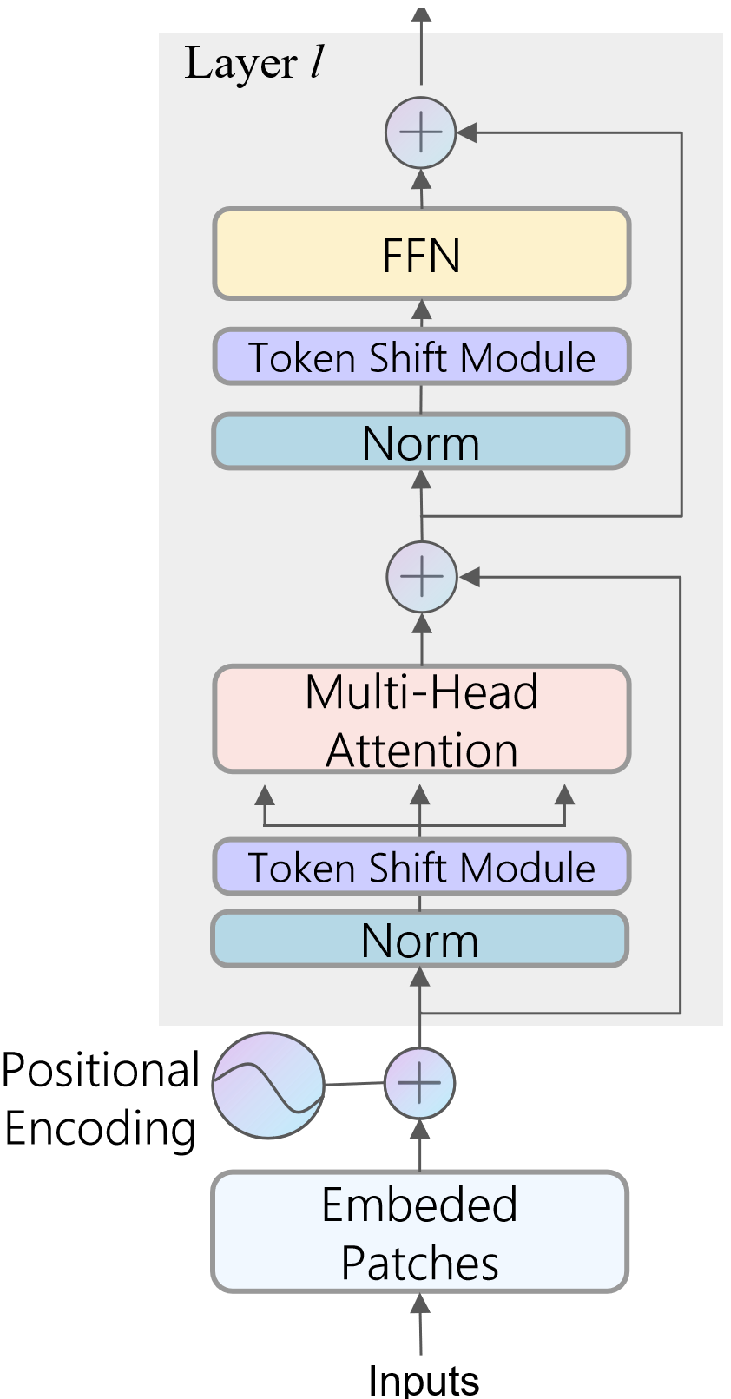}
        \caption{TokShift-\textit{B}}
        \label{fig:vist_b}
    \end{subfigure}
    \begin{subfigure}[t]{0.186\textwidth}
        \includegraphics[width=\textwidth]{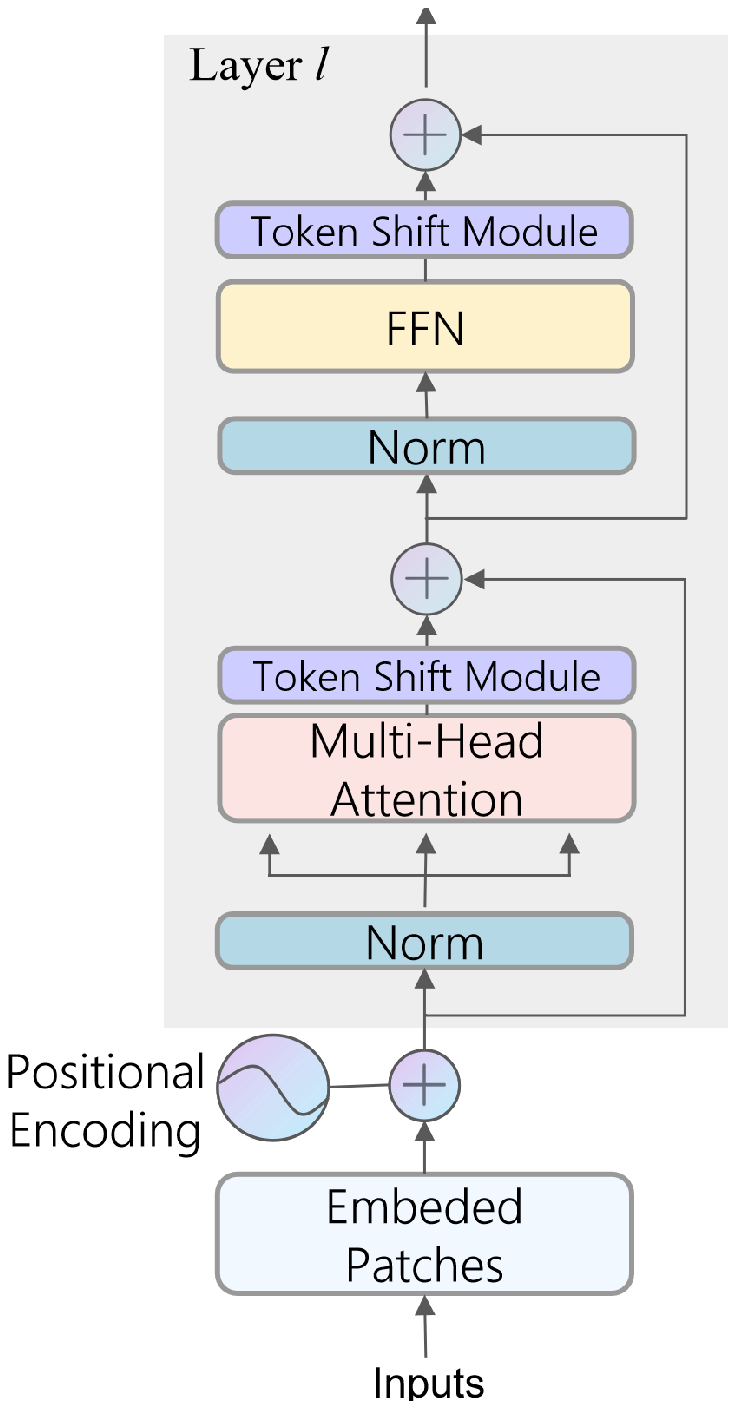}
        \caption{TokShift-\textit{C}}
        \label{fig:vist_c}
    \end{subfigure}

    \caption{\textbf{Integrations of TokShift module into ViT Encoder}: the TokShift can be plugged in multiple positions of a vision encoder: ``prior residual'' (TokShift), ``prior layer-norm'' (TokShift-\textit{A}), ``prior MSA/FFN'' (TokShift-\textit{B}), and ``post MSA/FFN (TokShift-\textit{C})''. Notably, a fronter position will affect the subsequent modules inside residual blocks more. }\label{fig:animals}
    \vspace{-0.3cm}
\end{figure*}
An overview of the Token Shift transformer (i.e., TokShift-xfmr) is presented in Figure (\ref{fig:overview}).~Our TokShift-xfmr follows a general pipeline of transformers, except for working on video inputs and containing extra temporal processing modules (i.e., the TokShift). 

To fit with transformer, a video $\pmb{v}\in \mathbb{R}^{T\times H\times W\times3}$ is firstly reshaped into sequential tensor $\pmb{\hat{v}}\in \mathbb{R}^{T\times N \times d}$, where $T$, $H/W$, $P$ separately represent clip-length, spatial-resolutions and patch-size, and $N=\frac{H\times W}{P^2}$, $d=P^2\times3$ denote the number of patches and the number of RGB pixels in a patch respectively. Then, each $i$-th patch $\pmb{x}^i_0$ in $\pmb{\hat{v}}$ is linearly projected into embedding space by $\pmb{E}$ and further added with the corresponding spatial positional embeddings $\pmb{E}_{pos}$. Similar as BERT \cite{devlin2018bert} and ViT \cite{dosovitskiy2020image}, an extra [\textit{Class}] token tensor $\pmb{c}_{0}\in \mathbb{R}^{T\times D}$ is concatenated with patches' embedding to represent per-frame global contents. Finally, the video $\pmb{\hat{v}}$ is embeded as $\pmb{z}_{0}$ (Equation (\ref{eq:vid}) and Figure (\ref{fig:tensor})).
\begin{align}
        &\pmb{z}_{0} = \left[ \pmb{c}_{0}; \pmb{x}^{1}_{0}\pmb{E};\pmb{x}^{2}_{0}\pmb{E};...;\pmb{x}^{i}_{0}\pmb{E};...;\pmb{x}^{N}_{0}\pmb{E}\right]+\pmb{E}_{pos} \label{eq:vid}\\
        &\pmb{x}^i_0\in\mathbb{R}^{T\times d},~~i=1,2,..., N \nonumber\\
        &\pmb{E}\in \mathbb{R}^{d\times D},~~\pmb{E}_{pos}\in \mathbb{R}^{(N+1)\times D} \nonumber\\
        &\pmb{z}_{0}\in \mathbb{R}^{T\times(N+1)\times D}\nonumber
\end{align}

The \textbf{TokShift-xfmr} (Figure (\ref{fig:overview})) contains $L$ replicated, identical encoders. An encoder consists of the TokShift module, Layer-Norm (LN), Multi-Head Self-attention (MSA), and Feed-Forward Network (FFN). The workflow for connecting them is shown by Figure (\ref{fig:overview}) or  Equations (\ref{eq:tokshift}-\ref{eq:msa} \& \ref{eq:tokshift2}-\ref{eq:zl}). We will describe each module separately as below.

The \textbf{TokShift module} (Equation (\ref{eq:tokshift}) \& (\ref{eq:tokshift2})) serves to in-placed manipulate global token $\pmb{c}/\pmb{\hat{c}}_{l-1}$ along temporal axis, introducing temporal interactions before feeding video embedding $\pmb{z}/\pmb{\hat{z}}_{l-1}$ through MSA/FFN. Hereby, the token $\pmb{c}/\pmb{\hat{c}}_{l-1}$ is the special [\textit{Class}] word of $\pmb{z}/\pmb{\hat{z}}_{l-1}$. We will present details of the TokShift and its variants in section \ref{sec3:tokshift}.  
\begin{align}
        &\pmb{c}_{l-1}=\textit{TokShift}\left( \pmb{c}_{l-1}\right)\label{eq:tokshift}\\
        &\pmb{z}_{l-1}=\textit{LN}\left(\pmb{z}_{l-1}\right)\\
        &\pmb{\hat{z}}_{l-1}=\textit{MSA}\left( \pmb{z}_{l-1}\right) + \pmb{z}_{l-1},\label{eq:msa}\\
        &\pmb{z}/\pmb{\hat{z}}_{l-1} \in \mathbb{R}^{T\times(N+1)\times D}\nonumber\\
        &\pmb{c}/\pmb{\hat{c}}_{l-1}=\mathbb{R}^{T\times D}\nonumber
\end{align}

The \textbf{MSA} (same for LN \& FFN) acts in the same way as in 2D ViT \cite{dosovitskiy2020image}. Specifically, its functions independently process each frame. Pick $\pmb{z}^{\left(t\right)}_{l-1}\in \mathbb{R}^{(N+1)\times D}$ at time-stamp $t$ of $\pmb{z}_{l-1}$ for example, functions of MSA are performed within each frame (Equations (\ref{eq:msa1})-(\ref{eq:msa3})).


\begin{align}
        &\textit{MSA}\left(\pmb{z}^{\left(t\right)}_{l-1}\right)=\textit{Concat}\left[\text{head}^{\left(t\right)}_{1}, \text{head}^{\left(t\right)}_{2}, ..., \text{head}^{\left(t\right)}_{M}\right]\label{eq:msa1}\\
        &\text{head}^{\left(t\right)}_{i}=\textit{Softmax}\left(\frac{\pmb{Q}^{\left(t\right)} \pmb{K}^{\left(t\right)}}{\sqrt{D}}\right)\pmb{V}^{\left(t\right)} \label{eq:msa2}\\
        &\pmb{Q}^{\left(t\right)}, \pmb{K}^{\left(t\right)}, \pmb{V}^{\left(t\right)}=\pmb{z}^{\left(t\right)}_{l-1}\times \pmb{W}_q, \pmb{W}_k, \pmb{W}_v\label{eq:msa3}\\
        &\pmb{W}_q, \pmb{W}_k, \pmb{W}_v \in \mathbb{R}^{D\times D} \nonumber
\end{align}

The \textbf{FFN} contains two linear projections and a GELU activation, serving to project each frame's feature independently.

\begin{align}
        &\pmb{\hat{c}}_{l-1}=\textit{TokShift}\left( \pmb{\hat{c}}_{l-1}\right)\label{eq:tokshift2}\\
        &\pmb{\hat{z}}_{l-1}=\textit{LN}\left(\pmb{\hat{z}}_{l-1}\right)\\
        &\pmb{z}_{l}=\textit{FFN}\left( \pmb{\hat{z}}_{l-1}\right) + \pmb{\hat{z}}_{l-1},~~~~~~~l=1,2,...,L\label{eq:zl}
\end{align}

Finally, a classification layer works on outputs (i.e., $\pmb{c}_{L}$) of the last encoder. Specifically, the video label is obtained by averaging frame-level predictions (Equation (\ref{eq:pred})), where \textit{FC} represents fully-connected layer of shape ``$D\times \text{Categories}$''. 
\begin{align}
        \pmb{y}&=\frac{1}{T}\sum_{i=1}^{T}\textit{FC}\left(\pmb{c}_L\right),~~~~~~~~\pmb{c}_{L}\in \mathbb{R}^{T\times D} \label{eq:pred}
\end{align}

\subsection{Shift Variants and the TokShift}\label{sec3:tokshift}
We propose and compare several shift variants, including the TokShift module, customized for the video transformer paradigm.

\textbf{Token Shift} merely manipulates [\textit{Class}] token tensors within each encoder (Figure~(\ref{fig:tensor_class})). Our motivation follows a principle that the current frame's partial contents are exchanged with prior/post time-stamps for dynamic motion while the rest are kept for static semantics. Coincidentally, the extra global [\textit{Class}] token tensor, aggregated from local features via weighted-sum (i.e., attention), is appropriate for implementing the principle.

Given a token tensor $\pmb{c}_{l}\in \mathbb{R}^{T\times D}$ representing the global sequence of a clip, the TokShift works as follows: $\pmb{c}_l$ is firstly split into 3 groups along channel dimension (Equation (\ref{eq:split})).
\begin{align}
        &\pmb{c}_{l}= \left[\pmb{s}_a, \pmb{s}_b, \pmb{s}_c\right]\label{eq:split}\\
        &\pmb{s}_a, \pmb{s}_b, \pmb{s}_c \in \mathbb{R}^{T\times a}, \mathbb{R}^{T\times b},\mathbb{R}^{T\times c}\nonumber\\
        &a+b+c=D\nonumber
\end{align}
Then, channels of splits $\pmb{s}_a$/$\pmb{s}_c$ are temporally shifted with prior/post time-stamps. For the split $\pmb{s}_b$, its content remains unchanged (Equations (\ref{eq:prior})-(\ref{eq:post})). The percentages of shifted channels are determined by $\frac{a}{D}$ \& $\frac{c}{D}$. 
\begin{align}
&\pmb{s}_{a}\left(t\right)=\pmb{s}_{a}\left(t-1\right)\label{eq:prior}\\
&\pmb{s}_{b}\left(t\right)=\pmb{s}_{b}\left(t\right)\\
&\pmb{s}_{c}\left(t\right)=\pmb{s}_{c}\left(t+1\right)\label{eq:post}\\
&t=1,2,..., T\nonumber
\end{align}

\textbf{Non Shift} module can be obtained by relpacing ``\textit{TokShift()}'' with ``\textit{Identity()}'' function in Equation~(\ref{eq:tokshift}) \& (\ref{eq:tokshift2}). This modification removes all temporal interactions and processes each frame independently. We name this simple 2D vision transformer for videos as ViT (video) for simplicity (i.e., Figure (\ref{fig:tensor}) \& (\ref{fig:vit})).

\textbf{Temporal Shift} extension for the encoder is obtained by imitating spatial/temporal shifts \cite{wu2018shift, you2020shiftaddnet,lin2019tsm} in CNN.~Recall that a CNN feature-map of image/video is a tensor collecting features from all the receptive fields (spatial or spatio-temporal locations). A shift operator then exchanges partial channels from neighboring receptive fields for each location on the feature-map. Similarly, in Figure (\ref{fig:tensor_shift}), temporal shift for encoder exchanges partial channels from prior/post time-stamps for all ``$N$ patches'' and ``$1$ [\textit{Class}] token''. 

\textbf{Patch Shift} is a variant of the temporal shift. Since the [\textit{Class}] token is a global representation, we exempt it from shift operation and only process patch embeddings (Figure (\ref{fig:tensor_patch})).

We compare the four shift variants on a standard benchmark and observe that the TokShift performs the best among them, showing the potential of capitalizing [\textit{Class}] tokens for temporal modeling in video transformers. We claim that TokShift introduces the minimum modifications on $\pmb{z}_l$ but grants the maximum benefits.

\subsection{Integration into ViT Encoder}\label{p3:position}
There are multiple candidate positions in an encoder to implant the TokShift module. The position is essential since it determines the degree of motions involved in representation learning. Specifically, from TokShift to TokShift-$A/B/C$ (Figure~(\ref{fig:vist_d})-(\ref{fig:vist_c})), the influence of motions decreases as fewer modules work upon shifted features.

A general encoder contains two residual blocks: one for self-attention, another for feature embedding (Figure~(\ref{fig:vit})). In each block, the TokShift can be placed at ``prior residual'', ``prior layer-norm'', ``prior MSA/FFN'' and ``post MSA/FFN''. We experimentally verify that placing TokShift at a front position (i.e., ``prior residual'') brings the best effect. This finding is different from TSM on CNN, where the optimal position is ``post residual". 

%% file: part4_exp.tex
\section{Experiments}
We conduct extensive experiments on three standard benchmarks for video classification and adopt Top-1/5 accuracy (\%) as evaluation metrics. We also report the model parameters and GFLOPs to quantify computations. 
\subsection{Datasets}
We adopt one large-scale (Kinetics-400) and two small-scale (EGTEA-Gaze+ \& UCF-101) datasets as evaluation benchmarks. Below presents their brief descriptions.

\textbf{Kinetics-400} \cite{carreira2017quo} serves as a standard large-scale benchmark for video classification. Its clips are truncated into 10 seconds duration with humans' annotations. The training/validation set separately contains $\sim$ 246k/20k clips covering 400 video categories. 

\textbf{EGTEA Gaze+} \cite{li2018eye} is a First-Person-View video dataset covering 106 fine-grained daily action categories. We select split-1 to evaluate our model, and this split contains 8,299/2,022 training/validating clips, with an average clip length of 3.1 seconds.

\textbf{UCF-101} \cite{soomro2012ucf101} contains 101 action categories, such as ``human-human/object'' interactions, sports and etc. We also select split-1 for performance reporting. This split contains 9,537/3,783 training/validating clips with average duration of 5.8 seconds.
\subsection{Implementations}
We implement a paradigm of shift-transformers, including the ViT (video), TemporalShift/PatchShift/TokShift-xfmr, on top of 2D ViT \cite{dosovitskiy2020image} with PyTorch. We also support backbones of various types or depths. 

All experiments share the same settings unless particularly specified. For ablations with higher resolutions, more frames, or heavier backbones, we follow the ``linear scaling rule'' \cite{goyal2017accurate} to adjust lr according to batch-size.

\textbf{Training}. Each clip, randomly cropped into 224$\times$224 (256/384 for high resolutions), contains 8/16 frames with a temporal step of 32. We apply training augmentations before cropping to reduce overfitting, including random resize (with the short side in [244, 330] and maintain aspect ratio), random brightness, saturation, gamma, hue, and horizontal-flip. We train the model for 18 epochs with batch-size 21 per GPU. Base lr is set to 0.1, is decayed by 0.1 at epoch [10, 15] during training. 
The \textit{Base-16} ViT \cite{dosovitskiy2020image} (contains 12 encoders) is adopted as the backbone and the shifted proportion is set to $\frac{a}{D}, \frac{c}{D}=\frac{1}{4}$. All experiments are run on $2/8\times$ V100 GPUs.

\textbf{Inference} adopts the same testing strategies as \cite{feichtenhofer2019slowfast}. Specifically, we uniformly sample 10 clips from a testing video. Each clip are resized to short-side=224 (256/384), then cropped into three 224$\times$224 sub-clips (from ``left'', ``center'', ``right'' parts). A video prediction is obtained by averaging scores of 30 sub-clips.

\subsection{Ablation Study}
Our ablation studies the impacts of the shift-types, integrations, hyperparameters of video transformers on the Kinetics-400 dataset. 

\textbf{Non Shift} \textit{vs} \textbf{TokShift}. We compare the Non Shift and TokShift under various sampling steps. Specifically, we fix clip size ($T$) to 8 frames, but change sampling steps ($S$) to cover variable temporal duration. 
\begin{table}[!htbp]
\footnotesize
\centering
\begin{tabular}{|l|l|l|l|l|}
    \hline
    \diagbox{Xfmr}{Acc1}{$T\times S$}&$8\times8$&$8\times16$&$8\times32$&$8\times64$\\ %
    \hline
    \hline
    ViT (video)\cite{dosovitskiy2020image}&76.17&76.32&76.02&75.73
    \\
    \textbf{TokShift}&76.90~(0.73$\uparrow$)&76.81~(0.49$\uparrow$)& \textbf{77.28}~(1.26$\uparrow$)&76.60~(0.87$\uparrow$)\\

    \hline
\end{tabular}
\caption{NonShift \textit{vs} TokShift under diffrent sampling strategies (``T/S'' refers to frames/sampling-step; Accuracy-1).}
\vspace{-0.6cm}
\label{tab:t1}
\end{table}

As in Tab. \ref{tab:t1}, the TokShift-xfmr outperforms ViT (video) under all settings. We observe that the performance of TokShift varies with different temporal steps. A smaller step makes TokShift not effective in learning temporal evolution. On the other hand, a large step of 64 frames results in a highly discontinuous frame sequence, affecting learning effectiveness. A temporal step of 32 frames appears to be a suitable choice.

\textbf{Integration}. We study the impacts of implanting the TokShift at different positions of an encoder. As mentioned in section \ref{p3:position}, a fronter position will affect the later modules inside residual blocks more. 
\begin{table}[h!t]
\begin{center}
\footnotesize
\begin{tabular}{|l|l|c|c|c|c|}
\hline
Model & Words Shiftted & $GFLOPs\times Views$ &Acc1&Acc5\\
 &  &  &(\%)&(\%)\\

\hline\hline
ViT (video) \cite{dosovitskiy2020image}& None &$134.7\times30$&76.02&92.52\\
\hline
TokShift-$A$& Token &$134.7\times 30$&76.85&93.10\\
TokShift-$B$& Token &$134.7\times 30$&77.21&92.81\\
TokShift-$C$& Token &$134.7\times 30$&77.00&92.92\\
\textbf{TokShift}& Token &$134.7\times 30$&\textbf{77.28} &92.91\\
\hline
\end{tabular}
\end{center}
\caption{Comparisons of the TokShift residing in different positions (GFLOPs, Accuracy-1/5).}
\vspace{-0.5cm}
\label{tab:t2}
\end{table}

Table \ref{tab:t2} lists comparisons of the TokShift residing in different positions. We observe that: (1). All models share the same GFLOPs, which is consistent with our zero-FLOPs claims on the TokShift; (2) Regardless of positions, the TokShift stably improves over the ViT (video) baseline; (3). Placing the TokShift at ``prior residual'', a front position that affects all later modules, presents the best top-1 accuracy.

\textbf{Words Shifted}. We assess contributions of the token-specific shift by comparing all shift variants in section \ref{sec3:tokshift}. 
\begin{table}[h!t]
\begin{center}
\footnotesize
\begin{tabular}{|l|l|c|c|c|c|}
\hline
Shift Type & Words Shiftted & $GFLOPs\times Views$ &Acc1&Acc5\\
 &  & &(\%)&(\%) \\
\hline\hline
ViT (video) \cite{dosovitskiy2020image}& None &$134.7\times 30$&76.02&92.52\\
\hline
TemporalShift & Token + Patches&$134.7\times 30$&72.88&91.24\\
PachShift &Patches                  &$134.7\times 30$&73.08&91.17\\ 
\textbf{TokShift}& Token       &$134.7\times 30$&\textbf{77.28} &\textbf{92.91}\\
\hline
\end{tabular}
\end{center}
\caption{Comparisons of shifting different visual ``Words'' (GFLOPs, Accuracy-1/5).}
\vspace{-0.5cm}
\label{tab:t3}
\end{table}

In Table \ref{tab:t3}, the TokShift exhibits the best top-1 accuracy. Additionally, consideration of patches in shift (TemporalShift \& PatchShift) reduces overall performance. The reason lies in that patches of the same spatial grid across times incurs visual misalignment when rigidly dividing a moving object. We verify this hypothesis by measuring the mean cosine similarity between the patch and its temporal neighbor features on kinetics-400 val set with Temporal/Patch Shift transformer and observe distances scores of 0.577/0.570. As for [\textit{Class}] token part, since it reflects global frame-level contents without alignment concerns, the distance between neighboring temporal tokens is 0.95. Consequently, for the TokShift or CNN-Shift, the drawback is bypassed by global alignment or sliding window.

\textbf{Proportion of shifted channels}. This hyperparameter controls the ratio of dynamic/static information in a video feature. We evaluate $\frac{a}{D}$,$\frac{c}{D}$ in range [${1}/{4}$, ${1}/{8}$, ${1}/{16}$] and present results in Table \ref{tab:tab4}. We experimentally find that ${1}/{4}$ is an optimal value, indicating that half channels are shifted (${1}/{4}$ back + ${1}/{4}$ forth) while the rest half remain unchanged.
\begin{table}[h!t]
\begin{center}
\footnotesize
\begin{tabular}{|l|c|c|c|}
\hline
Model & Channels Shifted & Acc1 &Acc5\\
& ($\frac{a}{D}+ \frac{c}{D}$) & (\%) &(\%)\\
\hline\hline
TokShift &${1}/{4}+{1}/{4}$& \textbf{77.28}&92.91\\ 
TokShift& ${1}/{8}+{1}/{8}$&77.18&92.95\\ 
TokShift& ${1}/{16}+{1}/{16}$&76.93&92.82\\ 
\hline
\end{tabular}
\end{center}
\caption{Comparisons of the TokShift regarding to proportions of shifted channels (Accuracy-1/5).}
\vspace{-0.5cm}
\label{tab:tab4}
\end{table}

\textbf{Word count} is an essential factor in transformer learning since more words indicate more details. As for videos, the number of visual words is positively correlated with two factors: spatial resolutions and temporal frames. We study the impacts of word count on the TokShift-xfmr. Table \ref{tab:tab5} lists the detailed performances under different word counts. Here, MR/HR represents middle/high resolutions.
\begin{table}[h!t]
\begin{center}
\footnotesize
\begin{tabular}{|l|c|c|c|c|c|}
\hline
Model & Res & Words/Frame & \#Frames& \# Words & Acc1\\
 &$H\times W$ & $N+1$ &$T$& $T\cdot (N+1)$&(\%) \\
\hline\hline
TokShift& $224\times224$ & $14^2+1$&6&1,182& 76.72\\
TokShift& $224\times224$ & $14^2+1$&8&1,576&77.28\\
TokShift& $224\times224$ & $14^2+1$&10&19,70&77.56\\
TokShift& $224\times224$ & $14^2+1$&16&3,152&\textbf{78.18}\\
\hline
TokShift (MR)& $256\times256$ & $16^2+1$&8&2,056&77.68\\
TokShift (HR)& $384\times384$ & $24^2+1$&8&4,616& 78.14\\
\hline
\end{tabular}
\end{center}
\caption{Comparisons of the TokShift according to visual word counts (Accuracy-1).}
\label{tab:tab5}
\vspace{-0.8cm}
\end{table}

We observe that: (1). Increasing word counts by either introducing more frames or spatial grids will lead to improvement (e.g., 76.72$\rightarrow$78.18 or 77.68$\rightarrow$78.14); (2). Increasing temporal resolution is more economic than spatial resolution. The former performs comparable (78.18 \textit{vs} 78.14) with fewer words than the latter (3,152 \textit{vs} 4,616).

\textbf{On various backbones}. We test the TokShift on transformers of various depths and types. Specifically, we further evaluate the TokShift-xfmr on \textit{Large-16} (24 encoders) and \textit{Hybrid-16} (12 encoders) ViTs. Compared to the Base-16, the Large-16 doubles the number of encoders. Hybrid-16 shares the same number of encoders but replaces the linear projection layer ($\pmb{E}$ in Equation (\ref{eq:vid})) with three residual blocks of ResNet50. Notably, for the TokShift-Large (HR), we set $T=12$ due to the memory limit. 

\begin{table}
\begin{center}
\footnotesize
\begin{tabular}{|l|l|c|c|c|}
\hline
Model &Backbone& Res & \# Words & Acc1 \\
 & & ($H\times W$) & $T\cdot(N+1)$ &(\%) \\
\hline\hline
TokShift (HR)&Base-16& $384\times384$ & $8\cdot(24^2+1)$&78.14\\
TokShift-Large (HR)&Large-16&$384\times384$ & $8\cdot(24^2+1)$&79.83\\
TokShift-Large (HR)&Large-16& $384\times384$ & $12\cdot(24^2+1)$&80.40\\
\hline
TokShift (MR)&Base-16& $256\times256$ & $8\cdot(16^2+1)$&77.68\\
TokShift-Hybrid (MR)&R50+Base-16&$256\times256$ & $8\cdot(16^2+1)$&77.55\\
TokShift-Hybrid (MR)&R50+Base-16&$256\times256$ & $16\cdot(16^2+1)$&78.34\\


\hline
\end{tabular}
\end{center}
\caption{Comparisons of the TokShift on different transformer backbones (Accuracy-1).}
\label{tab:tab6}
\vspace{-1.1cm}
\end{table}

Table \ref{tab:tab6} lists their comparisons with base backbones. The TokShift-xfmr performs better with deeper layers (Base-16 \textit{vs} Large-16). Whereas hybrid model achieves comparable performance as convolution free transformers, indicating that transformer is independent of convolutions. Additionally, performances further increase when using more frames (79.83$\rightarrow$80.40; 77.55$\rightarrow$78.34)

\begin{table*}[h!t]
\begin{center}
\scriptsize
\begin{tabular}{|l|c|c|c|c|c|c|c|c|}
\hline
Model &Backbone& Pretrain &Inference Res & \# Frames/Clip &GFLOPs$\times$Views & Params& Accuracy-1 & Accuracy-5 \\
 & & & ($H\times W$) &$T$&&&(\%)&(\%) \\
\hline\hline
I3D \cite{carreira2017quo} from \cite{feichtenhofer2020x3d} &InceptionV1&ImageNet& $224\times224$ & 250& $108\times$ NA & 12M&71.1 &90.3\\
Two-Stream I3D \cite{carreira2017quo} from \cite{feichtenhofer2020x3d}&InceptionV1&ImageNet& $224\times224$ & 500& $216\times$ NA & 25M&75.7 &92.0\\
S3D-G \cite{xie2018rethinking}&InceptionV1&ImageNet&$224\times224$&250&$71.3\times$NA&11.5M&74.7&93.4\\
Two-Stream S3D-G \cite{xie2018rethinking}&InceptionV1&ImageNet&$224\times224$&500&$142.6\times$NA&11.5M&77.2&93.0\\
Non-Local R50 \cite{wang2018non} from \cite{feichtenhofer2020x3d}&ResNet50&ImageNet& $256\times256$ & 128& $282\times 30$ & 35.3M&76.5 &92.6\\
Non-Local R101\cite{wang2018non} from \cite{feichtenhofer2020x3d}&ResNet101&ImageNet& $256\times256$ & 128& $359\times 30$ &54.3M&77.7 &93.3\\
\hline
TSM \cite{lin2019tsm} &ResNet50&ImageNet& $256\times256$ & 8& $33\times10$&24.3M &74.1&91.2\\
TSM \cite{lin2019tsm} &ResNet50&ImageNet& $256\times256$ & 16& $65\times10$&24.3M &74.7&-\\
TSM \cite{lin2019tsm} &ResNext101&ImageNet& $256\times256$ & 8& NA$\times10$&- &76.3&-\\
\hline
SlowFast $4\times16$ \cite{feichtenhofer2019slowfast}& ResNet50&None&$256\times256$ &32 & $36.1\times30$&34.4M&75.6&92.1\\
SlowFast $8\times8$ \cite{feichtenhofer2019slowfast}& ResNet50&None&$256\times256$ &32 &  $65.7\times30$&-&77.0&92.6\\
SlowFast $8\times8$ \cite{feichtenhofer2019slowfast}& ResNet101&None&$256\times256$ &32 & $106\times30$&53.7M&77.9&93.2\\
SlowFast $8\times8$ \cite{feichtenhofer2019slowfast}& ResNet101+NL&None&$256\times256$ &32 & $116\times30$&59.9M&78.7&93.5\\
SlowFast $16\times8$ \cite{feichtenhofer2019slowfast}& ResNet101+NL&None&$256\times256$ &32 & $234\times30$&59.9M&79.8&93.9\\
\hline
X3D-M \cite{feichtenhofer2020x3d}&X2D \cite{feichtenhofer2020x3d}&None&$256\times256$ & 16 &$6.2\times30$&3.8M&76.0&92.3\\
X3D-L \cite{feichtenhofer2020x3d}&X2D \cite{feichtenhofer2020x3d}&None&$356\times356$ & 16 &$24.8\times30$&6.1M&77.5&92.9\\
X3D-XL\cite{feichtenhofer2020x3d}&X2D \cite{feichtenhofer2020x3d}&None&$356\times356$ & 16 &$48.4\times30$&11M&79.1&93.9\\
X3D-XXL \cite{feichtenhofer2020x3d}&X2D \cite{feichtenhofer2020x3d}&None& NA &NA&$194.1\times30$&20.3M&\textbf{80.4}&94.6\\


\hline
ViT (Video)\cite{dosovitskiy2020image}&Base-16&ImageNet-21k& $224\times224$ & 8&$134.7\times30$ &85.9M &76.02&92.52 \\
\hline
\textbf{TokShift}&Base-16&ImageNet-21k& $224\times224$ & 8&$134.7\times30$ &85.9M& 77.28& 92.91\\
\textbf{TokShift} (MR)&Base-16&ImageNet-21k& $256\times256$ & 8&$175.8\times30$&85.9M& 77.68&93.55\\
\textbf{TokShift} (HR)&Base-16&ImageNet-21k& $384\times384$ & 8&$394.7\times30$ &85.9M&78.14&93.91\\
\textbf{TokShift}&Base-16&ImageNet-21k& $224\times224$ & 16&$269.5\times30$ &85.9M &78.18&93.78 \\
\textbf{TokShift-Large} (HR)&Large-16&ImageNet-21k& $384\times384$ & 8&$1397.6\times30$&303.4M&79.83&94.39\\
\textbf{TokShift-Large} (HR)&Large-16&ImageNet-21k& $384\times384$ & 12&$2096.4\times30$&303.4M&\textbf{80.40}&94.45\\

\hline
\end{tabular}
\end{center}
\caption{\textbf{Comparison to state-of-the-arts on Kinetics-400 Val}.}
\label{tab:all}
\vspace{-0.6cm}
\end{table*}

\subsection{Comparison with the State-of-the-Art}

We compare our TokShift-xfmrs with current SOTAs on Kinetics-400 datasets in Table \ref{tab:all}. Since there is no prior work that purely utilizes transformer, we mainly select 3D-CNNs, such as I3D, Non-Local, TSM, SlowFast, X3D as SOTAs. 

Compared to 3D-CNNs, transformers exhibit prominent performances. Particularly, the spatial-only transformer, i.e., ViT (video), achieves comparable or better performance (76.02) than strong 3D-CNNs, such as I3D (71.1), S3D-G (74.7) and TSM (76.3).  Moreover, a comparison of our most slimed TokShift-xfmr (77.28) with TSM (76.3) indicates that shift gains extra benefits on transformers than CNNs. Thirdly, the TokShift-xfmr (MR) achieves comparable performance (77.68) with Non-Local R101 networks (77.7) under the same spatial resolutions (255) but using fewer frames (8 \textit{vs} 128). This is reasonable since both contain the attentions. Finally, we compare TokShift-xfmr with the current best 3D-CNNs: SlowFast and X3D. The two 3D-CNNs are particularly optimized by a dual-paths or efficient network parameters searching mechanism. Since both SlowFast \& X3D introduce large structural changes over corresponding 2D nets, their authors prefer to train on videos with sufficient epochs (256), rather than initialize from ImageNet pre-trained weights. We show that our TokShift-Large-xfmr (80.40) performs better than SlowFast (79.80) and is comparable with X3D-XXL (80.4), where models are under their best settings. Notably, we only use 12 frames, whereas SlowFast/X3D-XXL uses 32/16+ frames.

In all, a pure transformer can perform comparable or better than 3D-CNNs for video classification. We additionally verify that temporal modeling can be imposed just on global [\textit{Class}] tokens rather than whole video embeddings. Besides, the common computational limitation of the transformer (appeared in NLP/vision task) also exists for videos. Though the TokShift introduces zero parameters \& FLOPs, the 2D transformer alone consumes more computations than 3D-CNNs. The reasons lie in: (1). Pair-wise distance calculation in attention is computationally expensive; (2). 3D-CNNs have been computationally optimized (TSM, P3D, etc.). Nevertheless, computation optimization for the transformer is already very hot for NLP \cite{choromanski2020rethinking,beltagy2020longformer} and image understanding \cite{touvron2020training}. As expected, computation optimization for long-length videos also requires future exploration.

\begin{figure*}
    \centering
    \begin{subfigure}[t]{0.38\textwidth}
       \includegraphics[width=\textwidth,height=3.5cm]{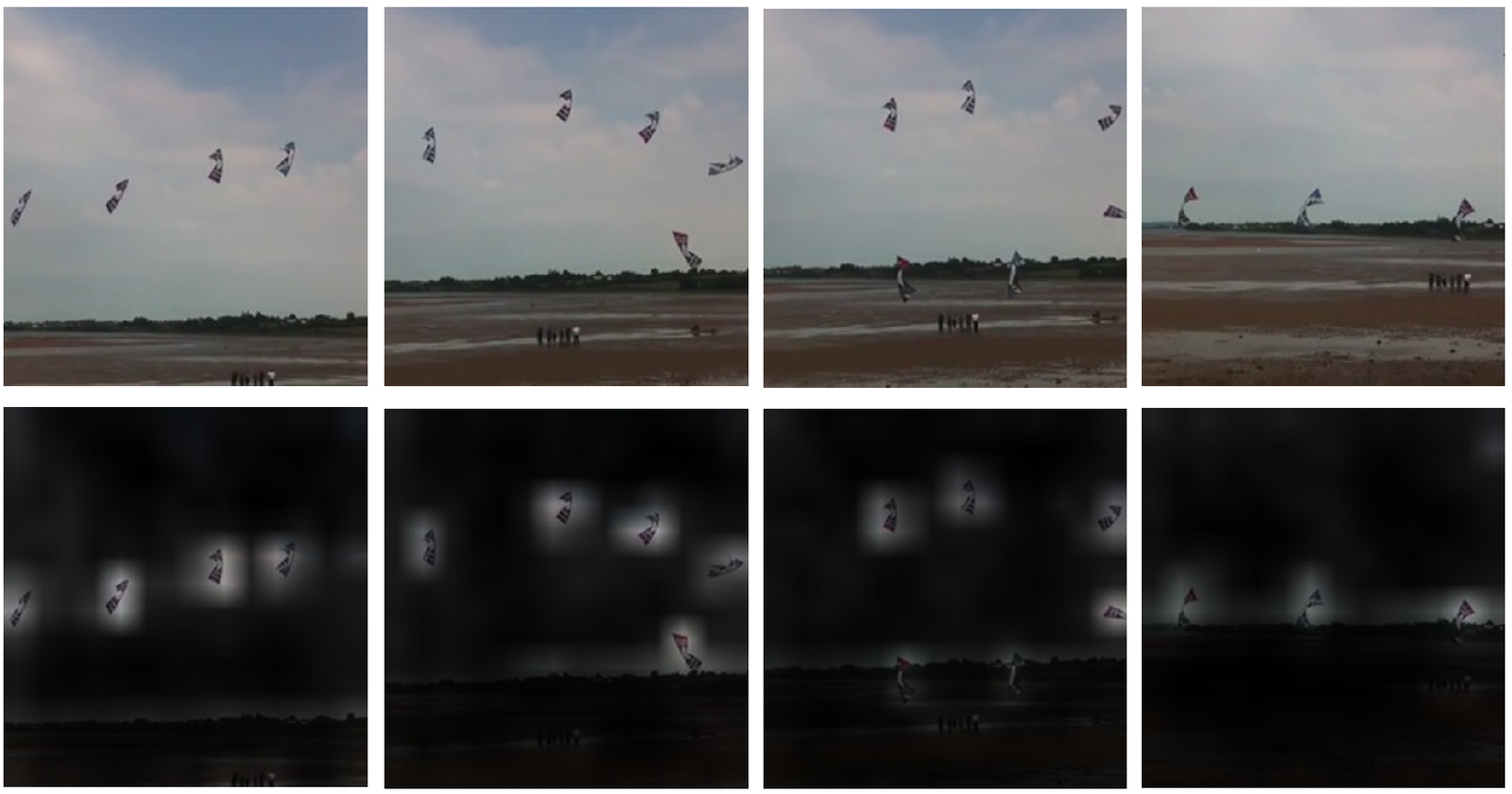} 
        \caption{``Flying kite''}
        \label{fig:vis55}
    \end{subfigure}
    \begin{subfigure}[t]{0.38\textwidth}
        \includegraphics[width=\textwidth, height=3.5cm]{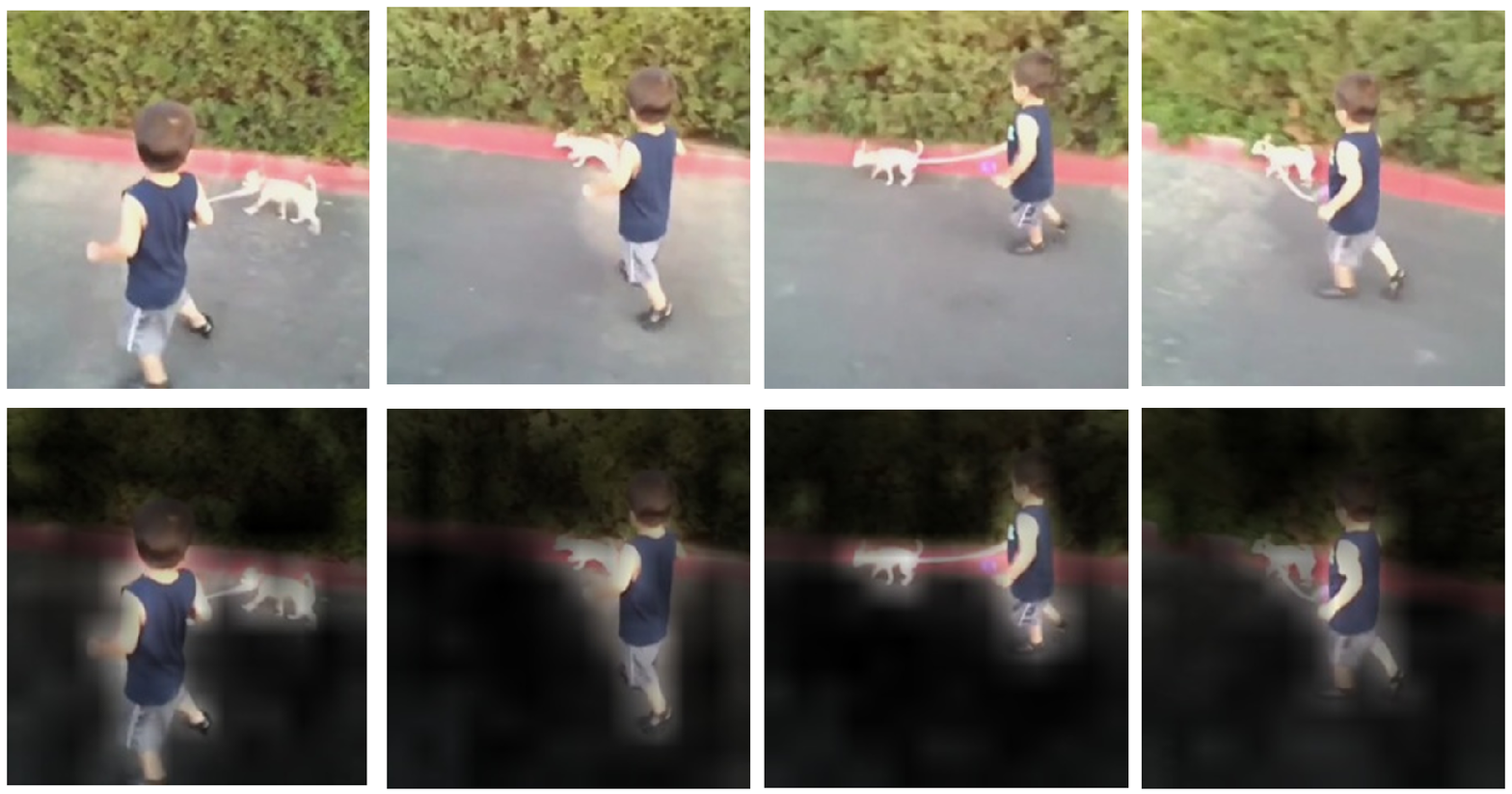}
        \caption{``Walking the dog''}
        \label{fig:vis22}
    \end{subfigure}
    \\
    ~ 
    \begin{subfigure}[t]{0.38\textwidth}
        \includegraphics[width=\textwidth,height=3.6cm]{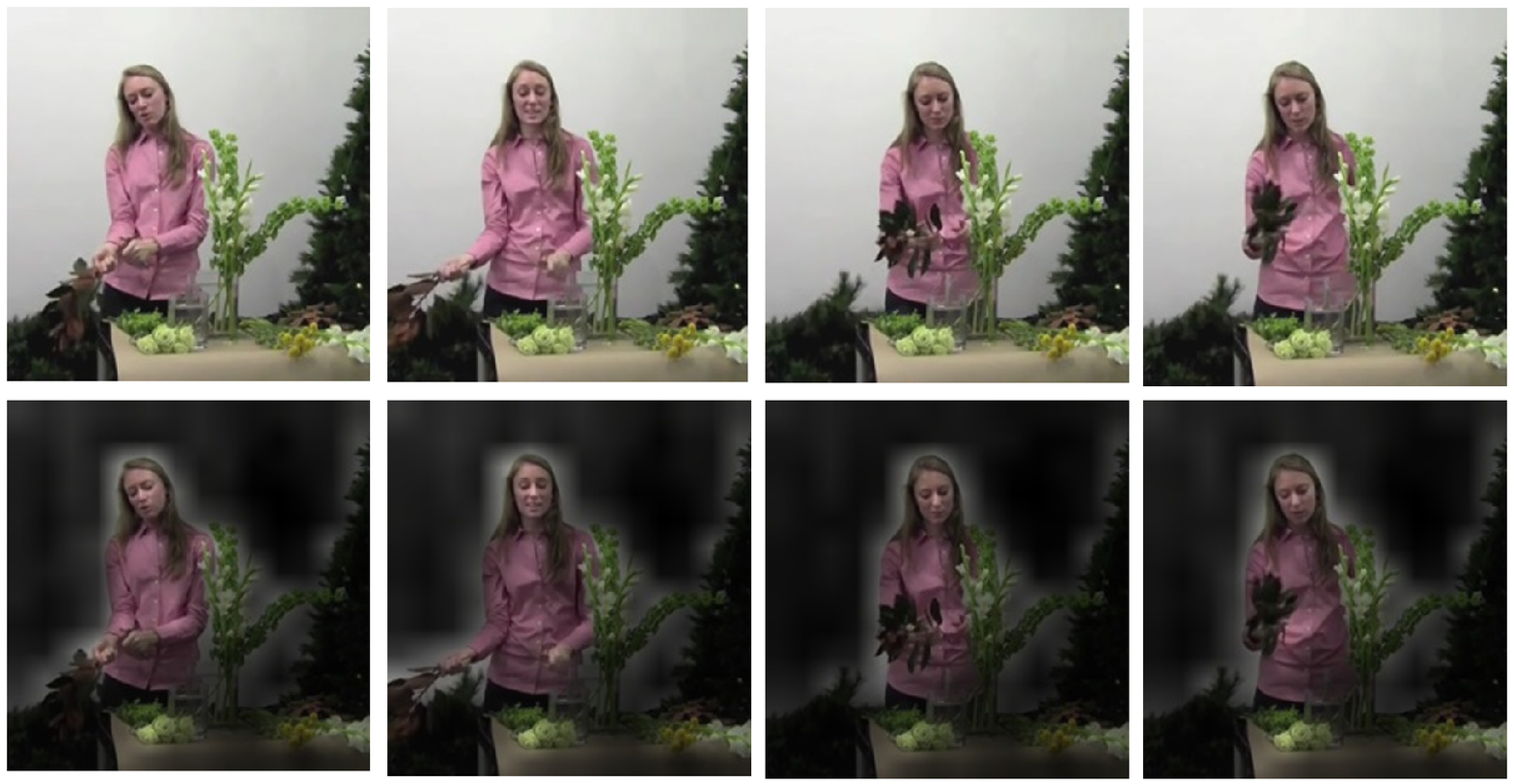}
        \caption{``Arranging flowers''}
        \label{fig:vis66}
    \end{subfigure}
    \begin{subfigure}[t]{0.38\textwidth}
        \includegraphics[width=\textwidth, height=3.6cm]{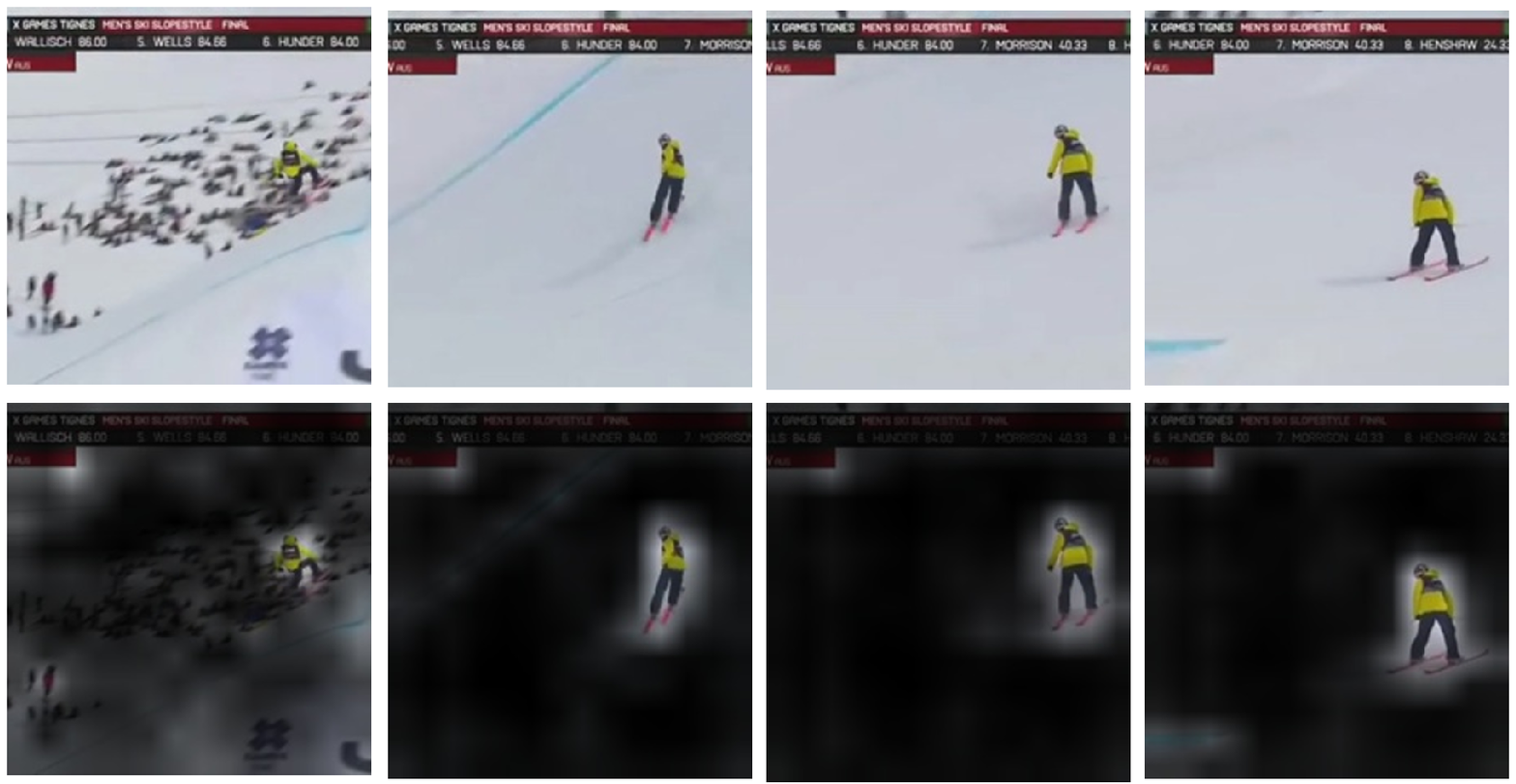}
        \caption{``Skiing''}
        \label{fig:vis11}
    \end{subfigure}
    \\
    ~ 
    \begin{subfigure}[t]{0.38\textwidth}
        \includegraphics[width=\textwidth,height=3.6cm]{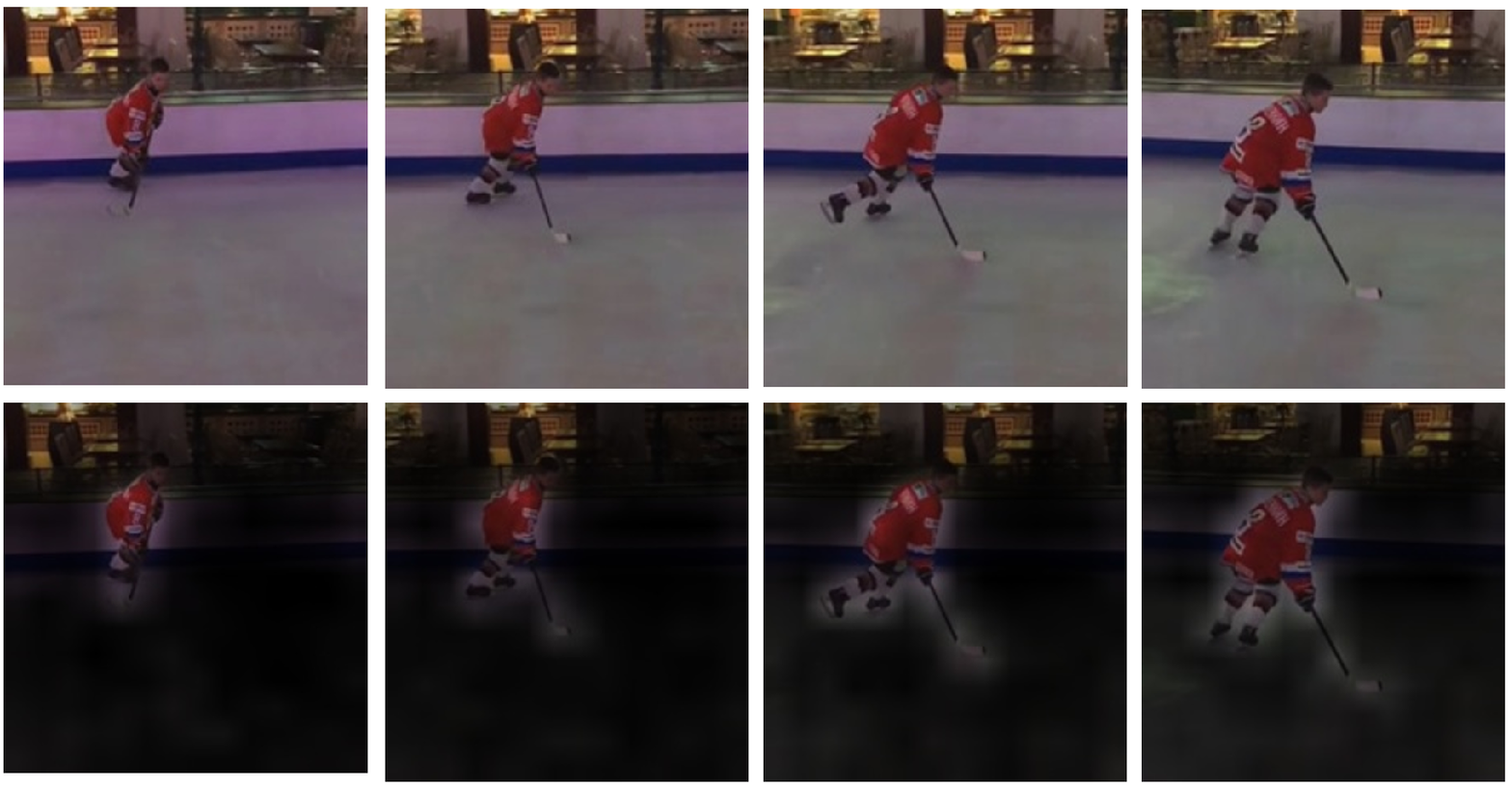}
        \caption{``Playing ice hockey''}
        \label{fig:vis44}
    \end{subfigure}
    \begin{subfigure}[t]{0.38\textwidth}
        \includegraphics[width=\textwidth,height=3.6cm]{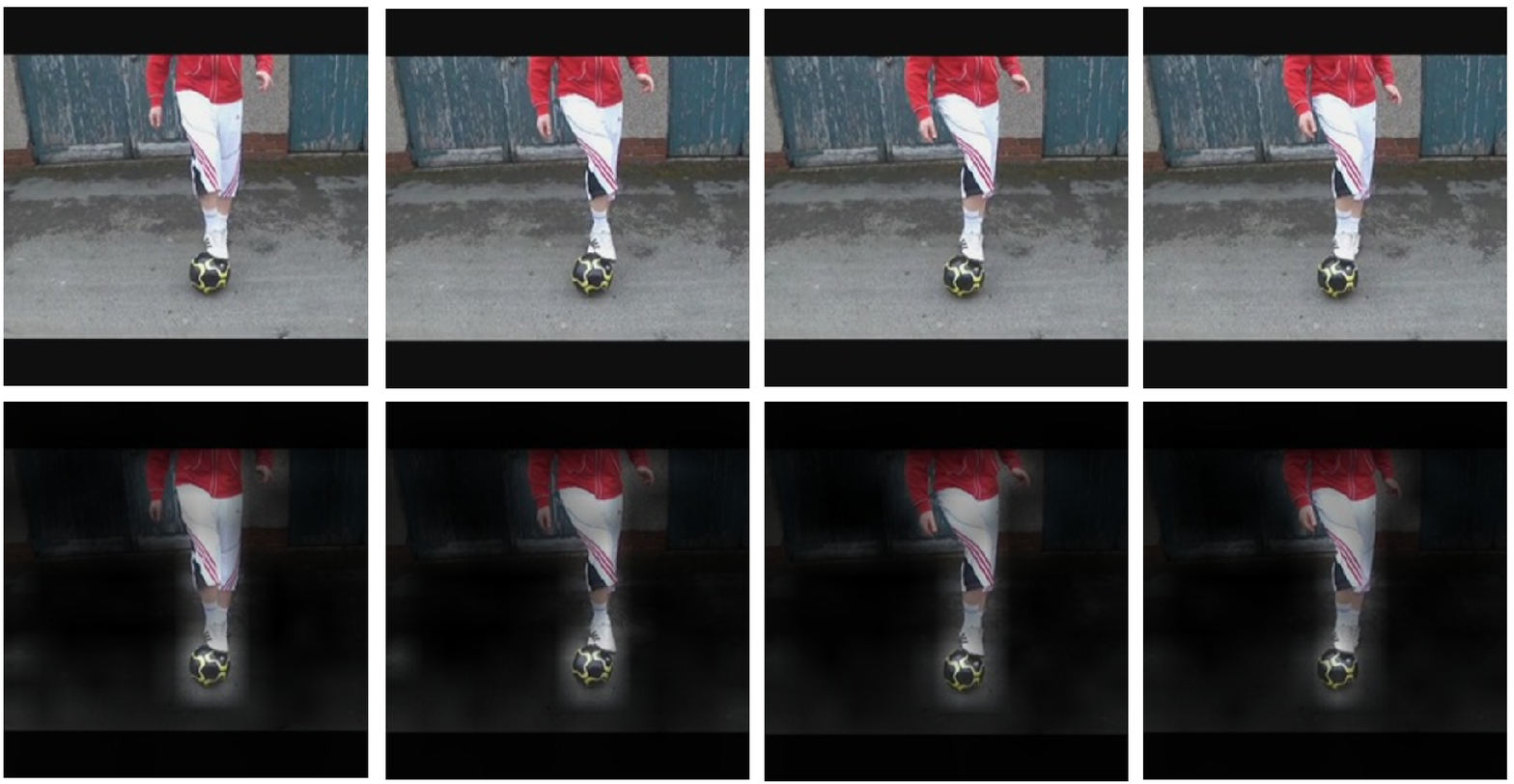}
        \caption{``Juggling soccer ball''}
        \label{fig:vis33}
    \end{subfigure}
    \caption{Visualization of attention on sample clips from the Kinetics-400 dataset. The odd row presents original frames, and the even presents corresponding attention maps.}\label{fig:attention}
\end{figure*}

\subsection{Attention Visualization}
An advantage of transformer over CNN is interpretability. Specifically, the self-attention emphasizes vital visual clues with high attentions. Hence, we can visualize the attention map to infer what the transformer values.

We adopt a public script\footnote{\url{https://github.com/jeonsworld/ViT-pytorch/blob/main/visualize_attention_map.ipynb}} to visualize the attention map from the last encoder. Maps from multiple heads are averaged to a heat map. We sample six clips, including daily and sports activities, from the Kinetics-400 and show their maps in Figure (\ref{fig:attention}). We keep half (4 of 8) frames for space-saving.  Notably, most action categories follow a form of ``verb'' + ``noun'' (e.g., ``flying kites''), meeting our intuitive explantations. Our TokShift-xfmr learns to value core parts such as ``kites'' in ``flying kites'', ``boy + dog'' in ``walking a dog'' and etc. Moreover, for ``flying kites'', the transformer even highlights all kites in the sky, indicating its potentials in counting applications.

\subsection{Fine-tune on small-scale datasets}
We study the impacts of pre-trained weights on small-scale datasets. Specifically, we fine-tune models on EGTEA Gaze+ and UCF-101 datasets with pre-weights from ``None'', ``ImageNet-21k'' and ``Kinetics-400''. The sampling step is reduced to 8, as both small datasets have a shorter duration (3/5s) than Kinetics-400 (10s). Training schemes are also optimized (EGTEA: 18 epochs, decay at [10,15]) and UCF: 25 epochs, decay at [10, 20]). 

\begin{figure}[h!t]
    \centering
    \scriptsize
    \begin{subfigure}[t]{0.13\textwidth}
        \includegraphics[width=\textwidth, height=\textwidth]{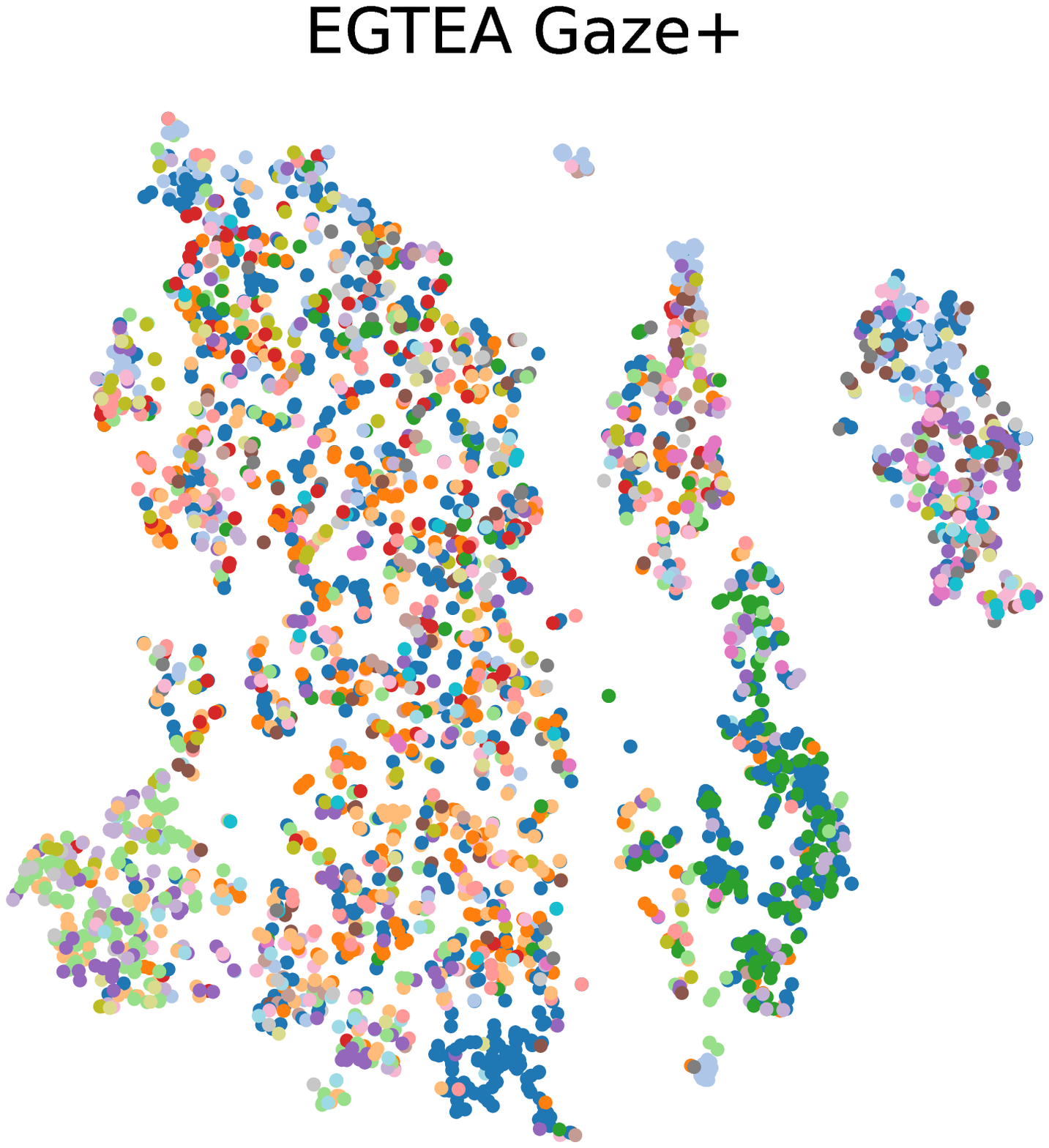}\\
        \vspace{-0.25cm}
        \caption{None}
        \label{fig:None}
    \end{subfigure}
    \begin{subfigure}[t]{0.13\textwidth}
        \includegraphics[width=\textwidth, height=\textwidth]{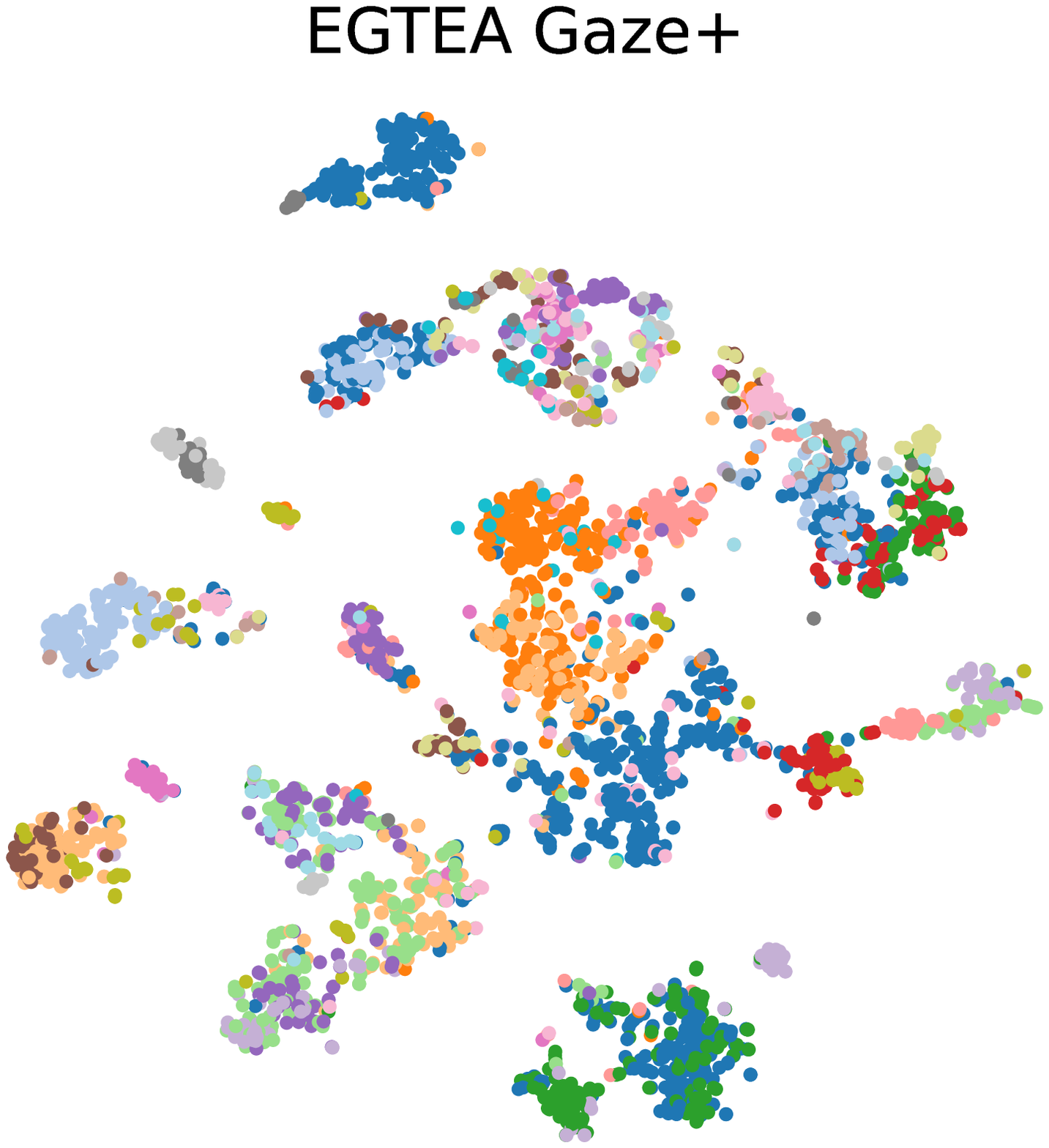}
        \caption{ImageNet-21k}
        \label{fig:Img21}
    \end{subfigure}
    \begin{subfigure}[t]{0.13\textwidth}
        \includegraphics[width=\textwidth, height=\textwidth]{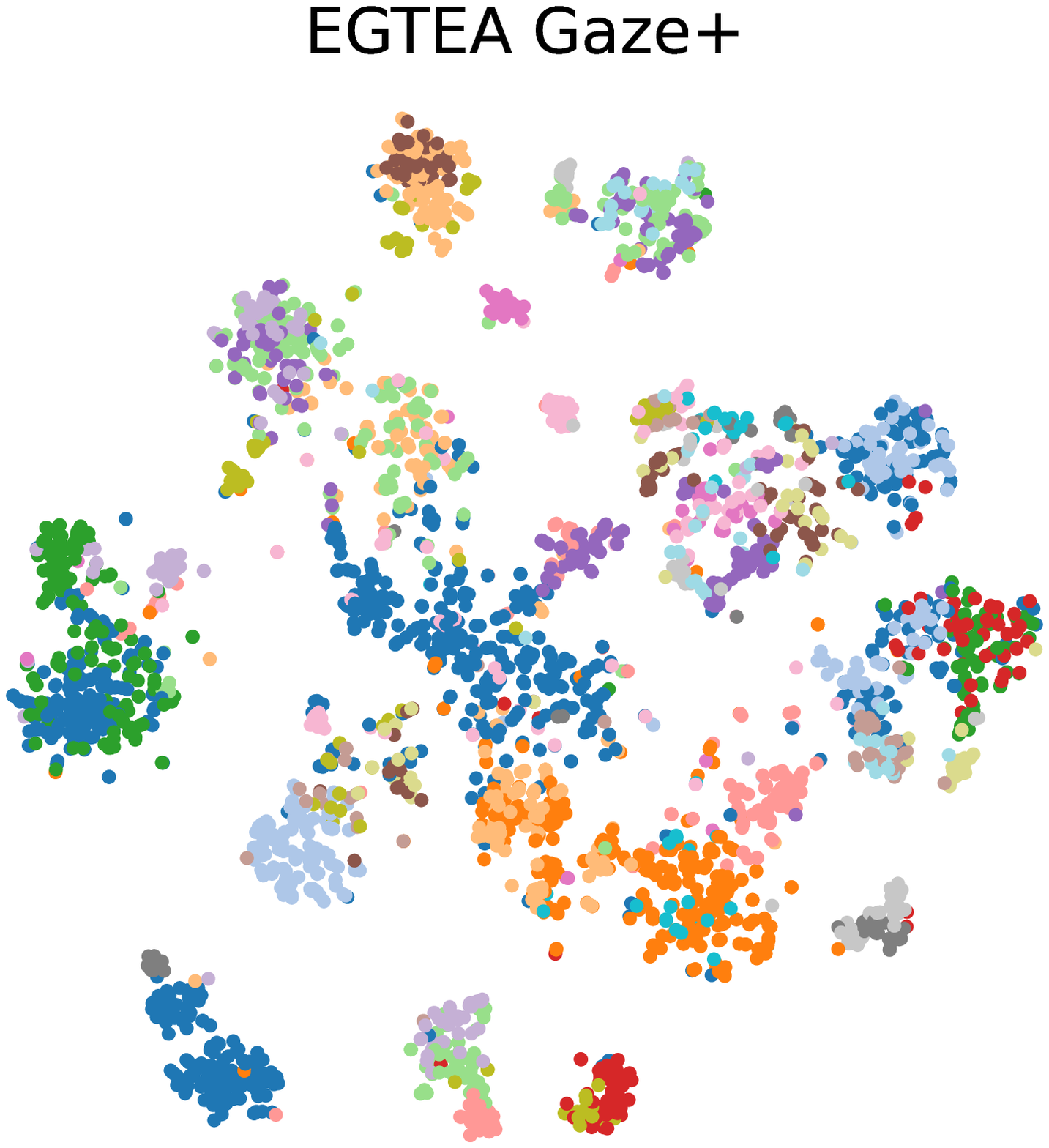}
        \caption{Kinetics-400}
    \label{fig:k400}
    \end{subfigure}
    \caption{Visualizing feature distributions with different pre-trained weights via t-SNE.}\label{fig:tsne}
    \vspace{-0.5cm}
\end{figure}

Table (\ref{tab:tab8}) \& (\ref{tab:tab9}) list their performances. Firstly, pre-training with Kinetics-400 outperforms the rest (i.e., \textbf{64.82} \textit{vs} 62.85 \textit{vs} 28.90 on EGTEA Gaze+; \textbf{95.35} \textit{vs} 91.65 \textit{vs} 91.60 on UCF-101). We project features of 60 categories from EGTEA Gaze+ with large improvements by t-SNE and find that the larger an upstream dataset is, the more scattered a downstream features are (Figure (\ref{fig:tsne})). Also, we study the impacts of freezing norm. Freezing ``layer-norms'' is beneficial for fine-tuning from Kinetics-400 weights, whereas harmful for ImageNet-21k weights. Because the Kinetics-400 shares the same modality (Video) with EGTEA Gaze+, while ImageNet-21k is different (Image). We verify the efficacy of two-stream TokShift-xfmr by adding the optical-flow (64.82$\rightarrow$65.08). Finally, we compare the TokShift with strong 3D-CNNs (I3D, P3D \& TSM), and verify its efficacy on small-scale datasets.

\begin{table}
\begin{center}
\scriptsize
\begin{tabular}{|l|l|l|c|c|c|}
\hline
Model & Modaility&Pretrain& Res & \# Frames & EGAZ+\\
 & && ($H\times W$) &$T$  & Acc1 (\%)\\

\hline\hline
TSM \cite{lin2019tsm}& RGB&Kinetics-400 & $224\times224$ &8 &63.45\\
SAP \cite{wang2020symbiotic} & RGB&Kinetics-400 & $256\times256$ &64 &64.10\\
ViT (Video) \cite{dosovitskiy2020image}&RGB&ImageNet-21k& $224\times224$ & 8 &62.59\\
\hline
TokShift& RGB&None& $224\times224$ & 8&28.90\\
TokShift &RGB&ImageNet-21k& $224\times224$ & 8&62.85\\
TokShift*&RGB&ImageNet-21k& $224\times224$ & 8&59.24\\
TokShift& RGB&Kinetics-400& $224\times224$ & 8&63.69\\
TokShift*&RGB&Kinetics-400& $224\times224$ & 8&\textbf{64.82}\\
TokShift& OptFlow&Kinetics-400& $224\times224$& 8&48.81\\
TokShift*-\textit{En}& OptFlow+RGB&Kinetics-400& $224\times224$ & 8&65.08\\
\hline
TokShift* (HR)& RGB&Kinetics-400& $384\times384$ & 8&65.77\\
TokShift-Large* (HR)& RGB&Kinetics-400& $384\times384$ & 8& \textbf{66.56}\\
\hline
\end{tabular}
\end{center}
\caption{Impacts of various pretrained weights on EGTEA-GAZE++ Split-1 dataset (``*'' means freeze layer-norm).}
\vspace{-0.6cm}
\label{tab:tab8}
\end{table}

\begin{table}
\begin{center}
\scriptsize
\begin{tabular}{|l|l|c|c|c|}
\hline
Model &Pretrain& Res & \# Frames &UCF101\\
 & & ($H\times W$) &$T$  & Acc1 (\%)\\

\hline\hline
I3D \cite{carreira2017quo}&Kinetics-400 & $224\times 224$ &250 &84.50\\
P3D \cite{qiu2017learning}&Kinetics-400 & $224\times 224$ &16 &84.20\\
Two-Stream I3D \cite{carreira2017quo}&Kinetics-400 & $224\times 224$ &500 &93.40\\
TSM \cite{lin2019tsm}&Kinetics-400 & $256\times 256$ &8 &95.90\\
ViT (Video) \cite{dosovitskiy2020image}&ImageNet-21k& $256\times256$ & 8 &91.46\\
\hline
TokShift&None& $256\times256$ & 8&91.60\\
TokShift &ImageNet-21k& $256\times256$ & 8&91.65\\
TokShift*&Kinetics-400& $256\times256$ & 8&95.35\\
\hline
TokShift* (HR)&Kinetics-400& $384\times384$ & 8&96.14\\
TokShift-Large* (HR)&Kinetics-400& $384\times384$ & 8&\textbf{96.80}\\
\hline
\end{tabular}
\end{center}
\caption{Impacts of various pretrained weights on UCF-101 Split dataset (``*'' means freeze layer-norm).}
\vspace{-0.6cm}
\label{tab:tab9}
\end{table}

